\documentclass{article}

\usepackage{fullpage}
\usepackage[utf8]{inputenc}
\usepackage[T1]{fontenc}
\usepackage{microtype}
\usepackage{indentfirst}

\usepackage{amsmath}
\usepackage{amsfonts}
\usepackage{amssymb}
\usepackage{amsthm}
\usepackage{mathtools}
\usepackage{bm}
\usepackage{commath}
\usepackage{relsize}
\usepackage{bbm}
\usepackage{nicefrac}
\usepackage{mysymbol}
\usepackage{wrapfig}

\usepackage{graphicx}
\usepackage{subfloat}
\usepackage{subcaption}
\usepackage{diagbox}
\usepackage{multirow}
\usepackage{booktabs}
\usepackage{hhline}
\usepackage{array}
\usepackage{pgfplots}
\usepackage{tikz}
\usetikzlibrary{positioning}

\usepackage{color}
\usepackage{tcolorbox}

\usepackage{algorithm}
\usepackage{algpseudocode}
\usepackage{listings}

\usepackage{cite}
\usepackage{url}

\usepackage{pifont}
\usepackage{newfloat}
\usepackage{enumerate}
\usepackage{enumitem}
\usepackage[title]{appendix}
\usepackage[font={small}]{caption}

\newtheorem{theorem}{Theorem}[section]
\newtheorem{proposition}[theorem]{Proposition}

\theoremstyle{definition}

\theoremstyle{remark}

\allowdisplaybreaks
\newcolumntype{?}{!{\vrule width 1pt}}

\makeatletter
\def\Let@{\def\\{\notag\math@cr}}
\makeatother

\usepackage[hidelinks,backref=page]{hyperref}
\definecolor{darkred}{RGB}{150,0,0}
\definecolor{darkgreen}{RGB}{0,150,0}
\definecolor{darkblue}{RGB}{0,0,150}
\hypersetup{colorlinks=true, linkcolor=darkred, citecolor=darkblue, urlcolor=darkblue}

\renewcommand*{\backref}[1]{}
\renewcommand*{\backrefalt}[4]{%
    \ifcase #1 (Not cited.)%
    \or        (Cited on page~#2.)%
    \else      (Cited on pages~#2.)%
    \fi}
    
\title{\textbf{Syndrome-Flow Consistency Model Achieves One-step Denoising Error Correction Codes}}

\date{}

\author{
Haoyu Lei\thanks{Department of Computer Science and Engineering, The Chinese University of Hong Kong, hylei22@cse.cuhk.edu.hk},
Chin Wa Lau\thanks{Huawei Technologies Co., Ltd., Theory Lab},
Kaiwen Zhou\thanks{Huawei Technologies Co., Ltd., Noah's Ark Lab}, 
Nian Guo\thanks{Huawei Technologies Co., Ltd., Theory Lab},
Farzan~Farnia\thanks{Department of Computer Science and Engineering, The Chinese University of Hong Kong, farnia@cse.cuhk.edu.hk}
	}

\begin{document}
\maketitle

\begin{abstract} 
Error Correction Codes (ECC) are fundamental to reliable digital communication, yet designing neural decoders that are both accurate and computationally efficient remains challenging. Recent denoising diffusion decoders achieve state-of-the-art performance, but their iterative sampling limits practicality in low-latency settings. To bridge this gap, consistency models (CMs) offer a potential path to high-fidelity one-step decoding. However, applying CMs to ECC presents a significant challenge: the discrete nature of error correction means the decoding trajectory is highly non-smooth, making it incompatible with a simple continuous timestep parameterization. To address this, we re-parameterize the reverse Probability Flow Ordinary Differential Equation (PF-ODE) by soft-syndrome condition, providing a smooth trajectory of signal corruption. Building on this, we propose the \emph{Error Correction Syndrome-Flow Consistency Model (ECCFM)}, a model-agnostic framework designed specifically for ECC task, ensuring the model learns a smooth trajectory from any noisy signal directly to the original codeword in a single step. Across multiple benchmarks, ECCFM attains lower bit-error-rate (BER) and frame-error-rate (FER) than transformer-based decoders, while delivering inference speeds 30x to 100x faster than iterative denoising diffusion decoders.
\end{abstract}

\section{Introduction}
Error Correction Codes (ECC) play a central role in modern digital communications and have been essential in a wide range of applications, including wireless communication and data storage. The core task of an ECC decoder is to recover a message from a received signal corrupted by noise during transmission. Recently, inspired by the great success of deep neural networks, neural network-based decoders were introduced, which are capable of improving the performance scores of conventional, problem-specific algorithms such as Belief Propagation (BP)~\cite{richardson2002capacity} and Min-Sum (MS)~\cite{fossorier1999reduced}.
\begin{figure}[t]
  \centering
  \includegraphics[width=0.66\textwidth]{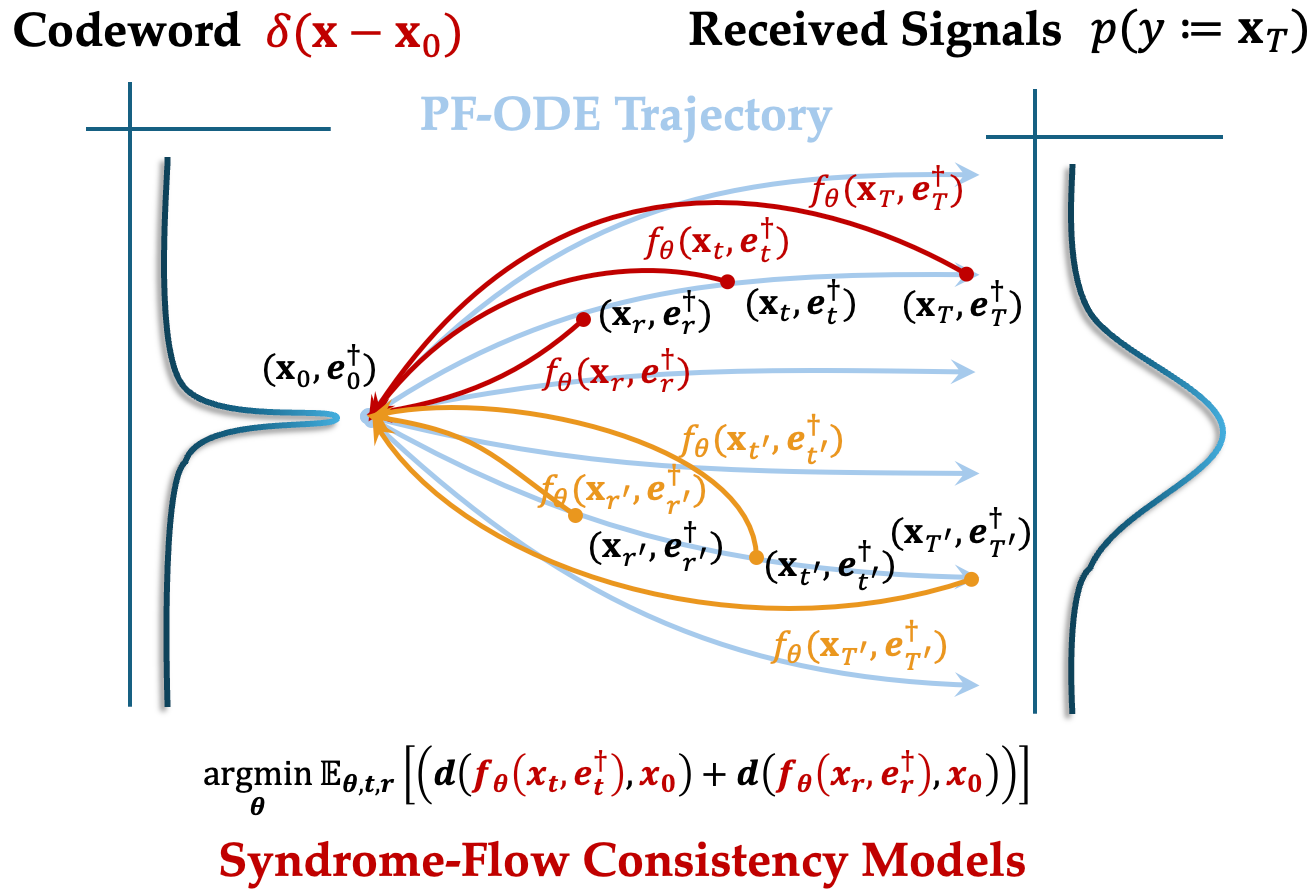}
  \caption{Illustration of our proposed Error Correction Syndrome-Flow Consistency Model (ECCFM). ECCFM learns to map received signals from the trajectories to a single, consistent codeword prediction $\mathbf{x}_0$, represented by a $\delta$ function.}
  \label{fig1}
\end{figure}

The existing neural decoders can be categorized into two groups. Early model-based approaches~\cite{lugosch2017neural,nachmani2019hyper,zhu2020learning} achieved successful results by integrating neural networks into conventional decoding algorithms. However, their reliance on problem-specific structures can limit the applicability as a general-purpose decoder. To address this limitation, model-free decoders enable the extension of a general neural network architecture without prior knowledge of decoding algorithms. While the early proposals applying fully connected neural net architectures~\cite{gruber2017deep,cammerer2017scaling,kim2018communication} could lead to overfitting, the preprocessing techniques that utilize magnitude and syndrome vectors~\cite{bennatan2018deep} have been effective in reducing the error. These improvements have led to efficient transformer-based decoders~\cite{choukroun2022error}, which leverage the self-attention mechanism to improve the numerical performance on short block codes. The transformer-based decoders have been improved in several recent works~\cite{choukroun2024foundation, choukroun2024learning}. Notably, the recent work by~\cite{parkcrossmpt} integrates cross-attention between magnitude and syndrome, resulting in the state-of-the-art performance in the decoding task.

In parallel, the remarkable success of diffusion generative models~\cite{ho2020denoising,song2020score} has inspired a new direction for neural decoder. Diffusion models train a noise estimator to gradually denoise noisy input data and reverse a forward noise manipulation process, generating high-quality codewords by an iterative denoising process. DDECC~\cite{choukroun2022denoising} first extends the denoising diffusion framework to ECC, naturally modeling the AWGN channel as a forward diffusion process. A time-dependent transformer learns to denoise the received signal iteratively, recovering the original codeword. This framework provides complementary gains over transformer-based methods, particularly for long codes and low SNR, due to its iterative refinement denoising process. 

However, the iterative denoising process inherently suffers from high computational complexity and latency, creating a significant bottleneck for real-time communication systems. While the iterative refinement is the key to high accuracy, it stands in direct conflict with the requirement for fast decoding in practical deployments. This challenge prompts the following question: Can we design a neural decoder that maintains the performance of diffusion-based decoders but with significantly reduced latency?



To solve this tension between accuracy and latency, the consistency model (CM) framework proposed by \cite{song2023consistency} for general applications provides a potential path. Designed to directly learn the reverse denoising trajectory into a single, efficient step, CMs could potentially offer a direct path to a high-fidelity, one-step decoder. However, our numerical results indicate that a direct application of existing CMs to ECC decoding would not yield the optimal performance, as detailed in Appendix~\ref{vanilla-train}. This outcome appears to arise from a conceptual incompatibility between the continuous nature of CMs and the discrete nature of the ECC task. As we discuss, the core assumption of a standard CM is the existence of a smooth trajectory along a Probability Flow ODE (PF-ODE). In ECC, the level of corruption of a valid codeword is measured by sum of syndrome errors~\cite{choukroundenoising}. The decoding path would not be a smooth flow but a series of jumps across a discrete space of error patterns, leading to suboptimal results.

Our main contribution is to resolve this incompatibility by re-parameterizing the decoding trajectory itself. Following the recent work by~\cite{lau2025interplaybeliefpropagationtransformer}, we introduce a soft-syndrome as a time condition instead of the standard syndrome, which provides a continuous and differentiable measure of the level of corruption of signals. By using the soft-syndrome condition to parameterize the reverse PF-ODE process, we build a smooth trajectory that is native to the ECC problem, mapping the noisy signal towards the manifold of clean codewords. Building on this, we propose the \emph{Error Correction Syndrome-Flow Consistency Model (ECCFM)}, a model-agnostic consistency framework for training single-step decoders. The ECCFM is trained with a well-designed consistency objective that maps any points on the same soft-syndrome-defined trajectory to the original clean codeword, as illustrated in Figure~\ref{fig1}. This forces the model to learn the decoding trajectory by a single consistency function.

We numerically evaluate the effectiveness of ECCFM through experiments on a diverse set of standard codes, including BCH, LDPC, Polar, MacKay, and CCSDS. The results show that ECCFM can improve bit-error-rate (BER) and frame-error-rate (FER) compared to leading transformer-based methods, exhibiting strong gains on longer codes with code lengths above $n=200$. Notably, it achieves an inference speedup of 30x to 100x over the existing iterative denoising diffusion methods in a single step, while maintaining comparable decoding performance. In our numerical evaluation, ECCFM offers a model-free training paradigm for building ECC decoders that achieves near state-of-the-art performance with low latency, and requires only a single-step inference suited for real-world communication systems.

\section{Related Works}
\textbf{Neural Network-based ECC Decoders.} Neural decoders are broadly categorized into model-based and model-free approaches. Model-based methods unfold and enhance conventional algorithms like Belief Propagation (BP)~\cite{richardson2002capacity} and Min-Sum (MS)~\cite{fossorier1999reduced} with learnable message-passing components. This paradigm, explored across various architectures~\cite{dai2021learning, kwak2022neural, kwak2023boosting, lugosch2017neural, nachmani2019hyper, nachmani2021autoregressive, marinkovic2010performance, zhu2020learning}, consistently surpasses its traditional counterparts in performance~\cite{matsumine2024recent}, but can be constrained by the underlying algorithm's structure and difficulties in capturing long-range dependencies.

In contrast, model-free decoders treat decoding as an end-to-end learning problem. Early fully-connected architectures~\cite{gruber2017deep, cammerer2017scaling, kim2018communication} suffered from overfitting, a challenge later mitigated by pre-processing techniques like magnitude-syndrome decomposition~\cite{bennatan2018deep}. The advent of Transformers~\cite{vaswani2017attention} led to ECCT~\cite{choukroun2022error}, which framed decoding as an auto-regressive sequence-to-sequence task. Subsequent works built upon this foundation to improve generalization (FECCT~\cite{choukroun2024foundation}), enable joint encoder-decoder training (DC-ECCT~\cite{choukroun2024learning}), and achieve superior performance among auto-regressive methods with cross-attention (CrossMPT~\cite{parkcrossmpt}). More recently, DDECC~\cite{choukroun2022denoising} introduced a diffusion-based approach, framing decoding as a denoising process and outperforming auto-regressive decoders.

\textbf{Diffusion Generative Models.} Diffusion generative models~\cite{sohl2015deep, ho2020denoising, song2020score, karras2022elucidating} have achieved state-of-the-art synthesis quality across diverse domains such as images~\cite{dhariwal2021diffusion, rombach2022high, podell2024sdxl}, video~\cite{he2022latent, blattmann2023align}, text~\cite{lou2023discrete, nie2025large}, and graphs~\cite{sun2023difusco, li2023t2t, lei2025boosting}. Their primary limitation, however, is the significant inference latency caused by the iterative denoising process, which has spurred the development of numerous accelerated sampling methods~\cite{song2020denoising, lu2022dpm}.

The diffusion framework is particularly well-suited for ECC, as the AWGN channel provides a forward process~\cite{choukroun2022denoising}. While diffusion-based decoders have set a new state-of-the-art in performance~\cite{choukroun2022denoising, parkcrossmpt}, they inherit the same critical issue of high computational cost. To resolve this accuracy-latency trade-off, we turn to Consistency Models (CMs)~\cite{song2023consistency, song2023improved, geng2024consistency}. CMs are a recent advance in diffusion model acceleration, designed to learn a direct mapping from any point on a denoising trajectory to its origin. This allows for high-fidelity, single-step generation, offering a potential path toward efficient and powerful ECC decoders.

\section{Preliminaries}
\textbf{Error Correction Codes.} 
In the error correction code setting, we consider a linear codebook $\mathcal{C}$, defined by a $k \times n$ generator matrix $G$ and an $(n-k) \times n$ parity-check matrix $H$. Note that these matrices satisfy $GH^\top = 0$ over the binary field $\mathbb{F}_2$. The encoder maps a message $m \in \{0,1\}^k$ to an $n$-bit codeword $x \in \mathcal{C} \subset \{0,1\}^n$ via the linear transformation $x=mG$. The codeword $x$ is then modulated using Binary Phase-Shift Keying (BPSK), where $0 \mapsto +1$ and $1 \mapsto -1$, resulting in the signal $x_s \in \{-1, +1\}^n$. We suppose this signal is transmitted over an Additive White Gaussian Noise (AWGN) channel. The received signal $y$ is given by:
$y = x_s + z$, where the noise vector $z$ is sampled from an isotropic Gaussian distribution, $z \sim \mathcal{N}(0, \sigma^2 I_n)$.

The objective of a decoder is to estimate the original codeword $\hat{x}$ from the noisy signal $y$. An essential tool for error detection is the syndrome, calculated from a hard-demodulated formulation of the received signal, $y_b$, performed as $y_b = \text{bin}(\text{sign}(y))$. Here $\text{sign}(y)$ is $+1$ for $y \ge 0$ and $-1$ otherwise, and the $\text{bin}$ function maps $\{-1, +1\}$ back to $\{1, 0\}$. The syndrome is then computed as $s = H y_b^\top$. An error is detected if the syndrome is a non-zero vector (i.e., $s(y) \neq 0$). Following the pre-processing technique proposed by~\cite{bennatan2018deep}, the input vector to the neural network is $[|y|,s(y)]$ with length $n+(n-k)$ to avoid overfitting.

\textbf{Consistency Models}. CMs~\cite{song2023consistency} built upon diffusion generative models and were introduced to overcome the inference computational costs by enabling fast, one-step generation. The core principle is that: any two points $(\mathbf{x}_t, t)$ and $(\mathbf{x}_{r}, r)$ on the same PF-ODE trajectory should map to the same origin point $\mathbf{x}_0$. CMs learn a function $f_{\theta}(\mathbf{x}_t, t)$ that directly estimates the \textit{trajectory} from noisy data to clean data with a single step:
\begin{equation}\label{cm}
    f_\theta(\mathbf{x}_t,t)=\mathbf{x}_0,
\end{equation}
The training objective of CMs is to enforce the self-consistency property across a discrete time steps. The continuous time interval $[0, T]$ is discretized into $N-1$ sub-intervals, defined by timesteps $1=t_1 < \dots < t_N=T$. The model is then trained to minimize the following loss, which enforces that the model's output remains consistent at adjacent points on the same PF-ODE trajectory:
\begin{equation}\label{standard-cm}
\mathcal{L}_{\text{Standard-CM}}(\theta):=\mathbb{E}[w(t)d\left(f_\theta(\mathbf{x}_{t},t),f_\theta(\mathbf{x}_{r},r)\right)].
\end{equation}
Here, $f_\theta$ is the consistency network being trained and $w(t)$ denotes the time schedule. We provide a detailed review of diffusion generative models and consistency models in the Appendix, Section \ref{cm_prelim}.

\section{Methodology}
\subsection{Syndrome-Flow Consistency Property}
\begin{figure*}[t] 
    \centering    
    \subfloat{ 
    \includegraphics[width=0.95\columnwidth]{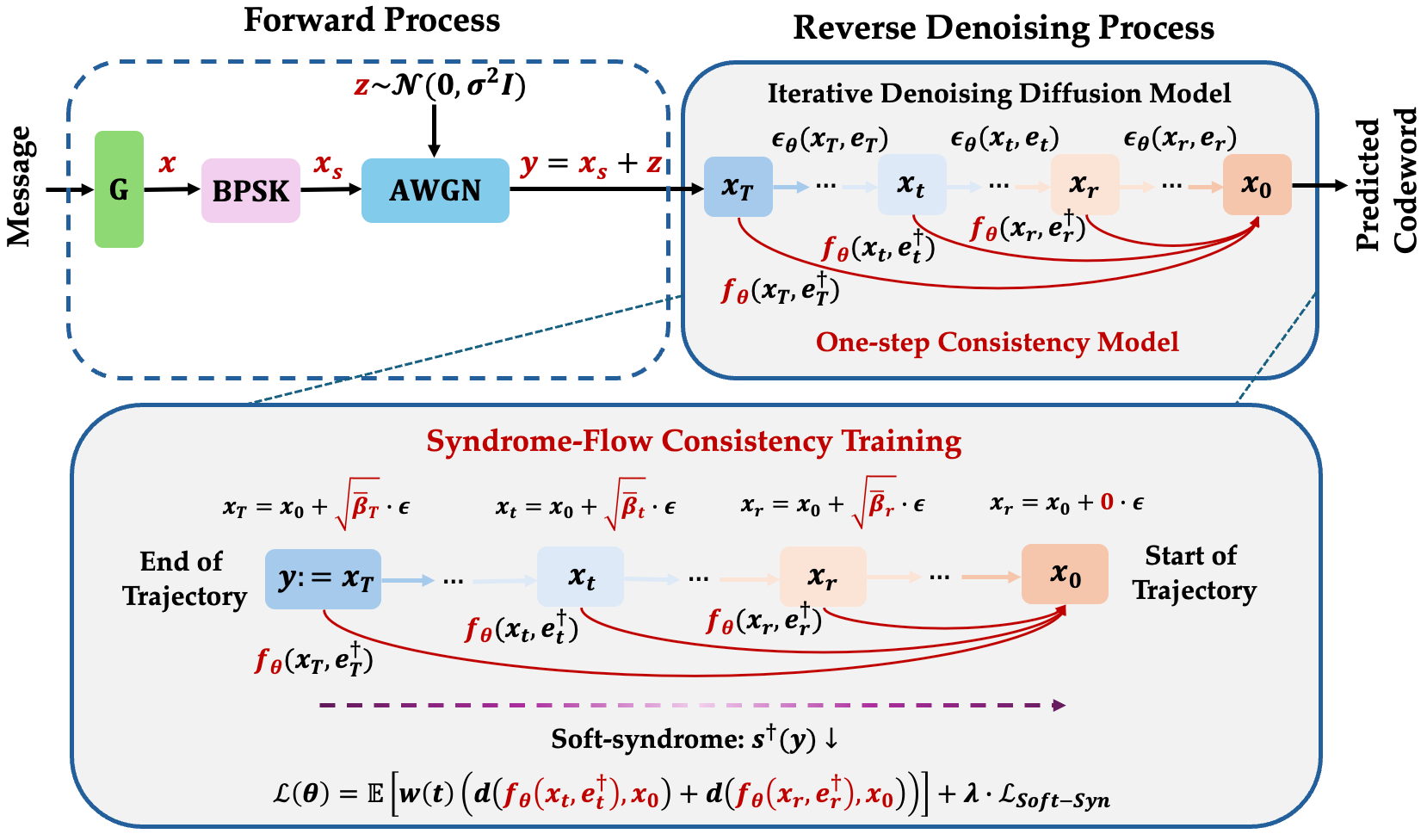}
    }
    \caption{Training Dynamics from iterative denoising to 1-step consistency decoding. DDECC's iterative diffusion denoising learns a noise predictor, $\epsilon_\theta(\cdot, e_t)$, requiring a multi-step iterative process to reverse the noise and decode the codeword. Our ECCFM, $f_\theta(\cdot, e^\dagger)$, directly learns the mapping from any noisy signal to the original clean codeword. By using the smooth soft-syndrome condition ($e^\dagger$), it achieves successful decoding in a single step.}
    \label{train}
    \vspace{-0.5cm}
\end{figure*}
While diffusion-based decoders like DDECC~\cite{choukroundenoising} achieve high performance, their iterative nature imposes significant computational overhead at inference. To enable fast, single-step decoding, we move beyond iterative refinement and propose a new paradigm based on the Syndrome-Flow Consistency, which defines the behavior of a one-step decoder. The consistency decoder should learn a function, $f_\theta$, that deterministically maps any noisy signals, $\mathbf{x}_t$, directly to its clean origin, $\mathbf{x}_0$, regardless of the noise level $t$. This concept is inspired by the self-consistency principles of Consistency Models (CMs)~\cite{song2023consistency, song2023improved}, but is specifically tailored to the deterministic nature of the error correction task. In decoding, the goal is not to sample from a distribution, but to recover a single, correct codeword defined by a Dirac delta function $\delta(\mathbf{x} - \mathbf{x}_0)$. Formally, we define the Syndrome-Flow Consistency property as follows:

\textbf{Syndrome-Flow Consistency Property.} Given a trajectory $\{y:=\mathbf{x}_t\}_{t\in[0,T]}$, we learn the consistency function as $f:(\mathbf{x}_t,t;\theta)\mapsto \mathbf{x}_0$, holding the Syndrome-Flow Consistency property: for a given ground-truth codeword $\mathbf{x}_0$, all points $(\mathbf{x}_t, t)$ on any trajectory originating from $\mathbf{x}_0$ map directly back to it, i.e., $f_\theta(\mathbf{x}_t, t) = \mathbf{x}_0, \quad \forall t \in [0, T]$. This implies that for any two noisy signals $\mathbf{x}_{t}$ and $\mathbf{x}_{r}$ derived from the same $\mathbf{x}_0$, their consistency function outputs must be identical and correct: $f_\theta(\mathbf{x}_{t}, t) = f_\theta(\mathbf{x}_{r}, r) = \mathbf{x}_0, \forall t,r\in[0,T]$. We learn the consistency model such that the conditional distribution is modeled as $p_\theta(\mathbf{x}|\mathbf{x}_t) = \delta(\mathbf{x}-f_\theta(\mathbf{x}_t, t))$, where $f_\theta(\mathbf{x}_t, t)$ directly estimates $\mathbf{x}_0$.

A naive application of the standard consistency training objective (Eq.~\ref{standard-cm}) is insufficient, as it only enforces relative consistency by minimizing $d(f_\theta(\mathbf{x}_{t},t), f_\theta(\mathbf{x}_{r},r))$. This is an indirect and inefficient objection for ECC task, where the ground truth $\mathbf{x}_0$ is known during training. We therefore introduce the Error Correction Consistency Loss (EC-CM), which directly enforces our proposed property by minimizing the distance between each predicted outcome and the true codeword $\mathbf{x}_0$:
\begin{align}\label{EC-CM}
\mathcal{L}_{\text{EC-CM}}(\theta)&:=\mathbb{E}[w(t)\left[d\left(f_\theta(\mathbf{x}_{t},t),\mathbf{x}_0\right)+d\left(f_\theta(\mathbf{x}_{r},r),\mathbf{x}_0\right)\right]],
\end{align}
In the ECC domain, we use Binary Cross Entropy (BCE) for the distance measure $d(\cdot, \cdot)$. Our Proposition~\ref{prop} theoretically demonstrates that our optimization objective, $\mathcal{L}_{\text{EC-CM}}$, also serves as an upper bound on the standard consistency loss measured by Total Variation distance. This confirms that our more direct approach inherently enforces self-consistency.

\begin{proposition}\label{prop}
Let $\mathcal{L}_{\text{Standard-CM}}$ be defined by the Total Variation distance, $\text{TV}(\cdot,\cdot)$, and $\mathcal{L}_{\text{EC-CM}}$ be defined by the Binary Cross Entropy, $\text{BCE}(\cdot,\cdot)$. For any timesteps $t, r$ and ground truth $\mathbf{x}0$, the following semi–triangle inequality holds:
\begin{equation}
\begin{aligned}
\text{TV}^{2}\bigl(f_\theta(\mathbf{x}_{t},t), f_\theta(\mathbf{x}_{r},r)\bigr)
&\le \text{BCE}\bigl(f_\theta(\mathbf{x}_{t},t),\mathbf{x}_{0}\bigr)
+\text{BCE}\bigl(f_\theta(\mathbf{x}_{r},r),\mathbf{x}_{0}\bigr).
\end{aligned}
\end{equation}
\end{proposition}
This  change in the training objective compels all decoding trajectories to map directly to their origin This fundamental change requires all decoding trajectories mapping to their origin codeword, as demonstrated in Figure~\ref{fig1}, and the learned solution distribution is expected to center on $\mathbf{x}_0$.

\subsection{Soft-syndrome Condition for Smooth Trajectory}

\begin{figure}[t]
  \centering
  \includegraphics[width=0.6\textwidth]{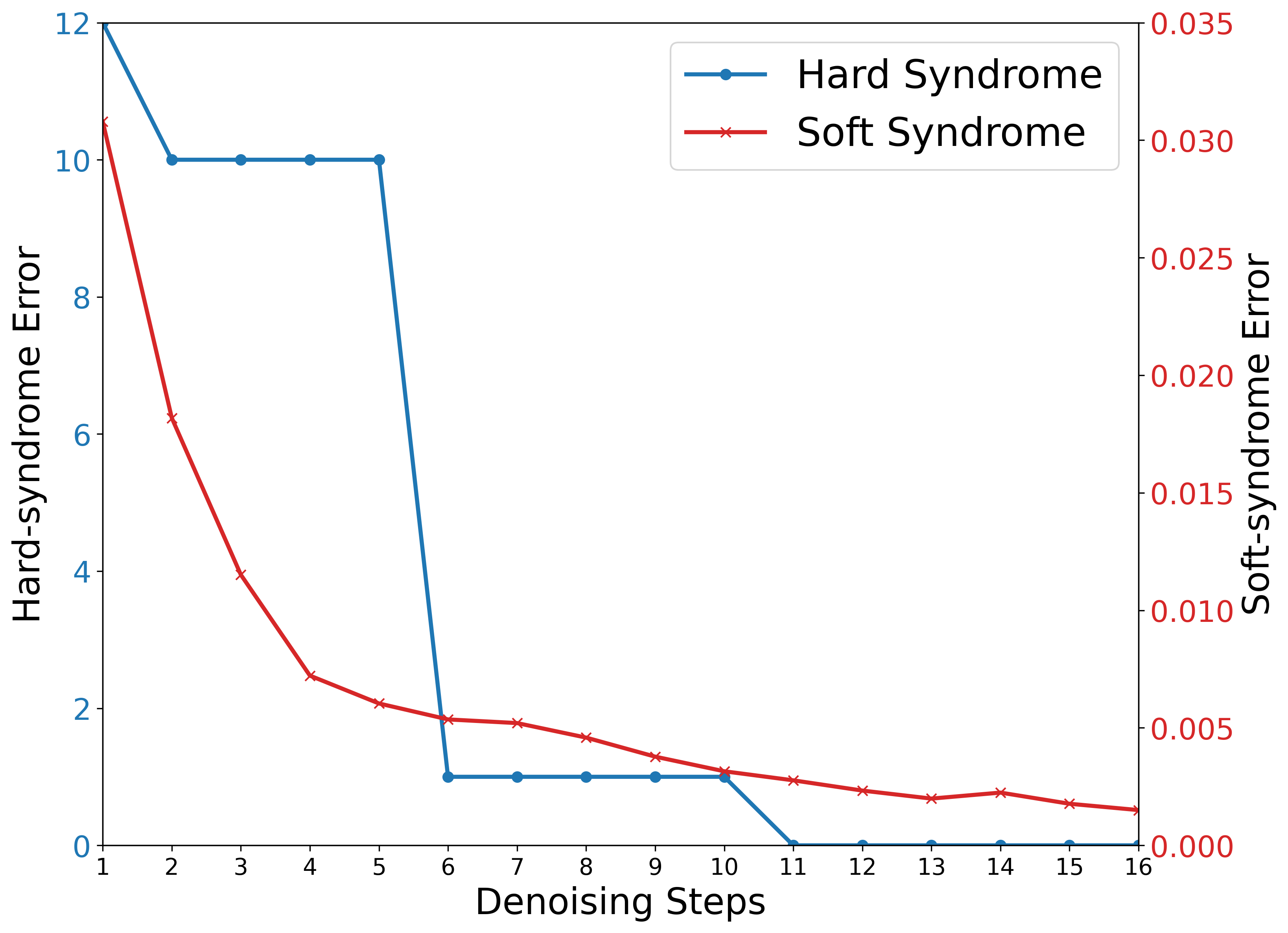}
    \caption{Decoding trajectories for models conditioned on Hard Syndrome versus Soft Syndrome on a POLAR(64,48) code. The soft-syndrome conditioning results in a smoother path to a valid codeword. Additional Results under low SNR are available in Appendix~\ref{abl_snr}, Figure~\ref{fig:syn_images}.}
  \label{traj}
\end{figure}

The training objective of a consistency model is rooted in the differential equation $\frac{\mathrm{d}f}{\mathrm{d}t}=0$. Following Eq. 10 and 11 in~\cite{geng2024consistency}, the consistency function $f_\theta$ is parameterized to satisfy these conditions:
\begin{equation}
f_\theta(\mathbf{x}_t,t)=\mathbf{x}_0\Leftrightarrow \frac{\mathrm{d}f}{\mathrm{d}t}=0,f_\theta(\mathbf{x}_0,0)=\mathbf{x}_0,
\end{equation}
In practice, this differential form is discretized for training using a \textit{finite-difference approximation}, by dividing the time horizon into $N-1$ sub-intervals $1=t_1<\cdot\cdot\cdot<t_{N}=T$:
\begin{equation}
0=\frac{\mathrm{d}f}{\mathrm{d}t}\approx\frac{f_\theta(\mathbf{x}_t,t)-f_\theta(\mathbf{x}_r,r)}{t-r},
\label{eq:finite_diff}
\end{equation}
where $\mathrm{d}t=\Delta t=t-r$, $t>r\geq0$.

A critical problem arises when applying the consistency framework to decoding tasks: the time variables $t$ and $r$, which measure noise levels, are not directly observable from the received signals. A natural solution, proposed in DDECC~\cite{choukroundenoising}, is to use the sum of syndrome error, $e_t$, as a measurement of the noise level, since $e_t=0$ indicates a valid codeword. The hard syndrome is computed as $s = Hy_b^\top$, where $H$ is the parity-check matrix. The sum of syndrome error, $e_t = \sum_i s(\mathbf{x}_t)_i$, is then the sum of binary syndrome bits. 

However, replacing the continuous time variables $t$ and $r$ in Eq.~\ref{eq:finite_diff} with their discrete, integer-valued counterparts $e_t$ and $e_r$ violates the core assumption of a smooth, differential trajectory in consistency training. As we demonstrate in Figure~\ref{traj}, the trajectory defined by the syndrome error is highly non-smooth and clustered. A small change in the noisy signal can cause an abrupt jump in the syndrome error count, invalidating the finite-difference approximation and leading to instability during training in~\ref{traj}.

Therefore, the discrete and non-smooth nature of the syndrome error makes it an unsuitable conditioning variable for consistency training in ECC, which necessitates a smooth and differential measure of noise level along the denoising trajectory. We then propose replacing this discrete error sum condition with a continuous and differentiable alternative. Inspired by~\cite{lau2025interplaybeliefpropagationtransformer}, we introduce the \textit{soft syndrome} as the basis for the consistency noise condition. The soft syndrome, $s^\dagger$, is a fully differentiable function that leverages the log-likelihood ratios of the received signal $\mathbf{x}_t$: and offers a continuous measure of how close each parity-check equation is to being fulfilled.

\begin{algorithm}[t]
\caption{Syndrome-Flow Consistency Training}
\label{alg:training}
\begin{algorithmic}[1]
    \Require Model $f_{\boldsymbol{\theta}}$, parity-check matrix $\mathbf{H}$, learning rate $\eta$, syndrome weight $\lambda$, denoising steps $N$, time scaling factor $\alpha$.
    
    \For{training batch $\mathbf{x}_{0}$}
        \State $t\sim \mathcal{U}\{1, \dots, N\},\,r=\alpha t$
        \Comment{Sample forward diffusion steps}
        
        \State $\boldsymbol{\epsilon}\sim \mathcal{N}(\mathbf{0}, \mathbf{I})$
        \Comment{Sample Gaussian noise}
        
        \State $\mathbf{x}_{t} \gets \mathbf{x}_{0} + \sqrt{\bar{\beta}_{t}} \cdot \boldsymbol{\epsilon},\mathbf{x}_{r} \gets \mathbf{x}_{0} + \sqrt{\bar{\beta}_{r}} \cdot \boldsymbol{\epsilon}$
        \Comment{Generate noisy signals via forward process}
        
        \State $e^\dagger_t=\mathcal{L}_{\text{Soft-syn}}(\mathbf{x}_t,H),\,\,e^\dagger_r=\mathcal{L}_{\text{Soft-syn}}(\mathbf{x}_r,H)$
        \Comment{Calculate soft-syndrome}
        
        \State $\mathcal{L}_{\text{EC-CM}} \gets \text{BCE}(f_\theta(\mathbf{x}_{t}, e^\dagger_t), \mathbf{x}_{\text{0}}) + \text{BCE}(f_\theta(\mathbf{x}_{r}, e^\dagger_r), \mathbf{x}_{\text{0}})$
        
        \State $\mathcal{L}_{\text{Total}} \gets \mathcal{L}_{\text{Consistency}} + \lambda \cdot \left(\mathcal{L}_{\text{Soft-syn}}(f_\theta(\mathbf{x}_{t}, e^\dagger_t),H)+\mathcal{L}_{\text{Soft-syn}}(f_\theta(\mathbf{x}_{r}, e^\dagger_r),H)\right)$
        
        \State $\boldsymbol{\theta} \gets \boldsymbol{\theta} - \eta\cdot \nabla_{\boldsymbol{\theta}} \mathcal{L}_{\text{Total}}$ 
        
    \EndFor
\end{algorithmic}
\end{algorithm}
\textbf{Differential Soft-syndrome Error Time Condition.} Similar to the conventional hard sum of syndrome error, we compute the soft-syndrome error condition $e^\dagger_t$ of the parity-check matrix $H$ as:
\begin{equation}
\begin{aligned}\label{soft-syn}
e^\dagger_t&:= -\frac{1}{n-k}\sum_{j}\log \Pr(S_j^\dagger = 0) = -\frac{1}{n-k}\sum_{j}\log (1 - s_j^\dagger),
\end{aligned}
\end{equation}
This soft-syndrome error condition is computed as the binary cross-entropy between the estimated syndrome and the all-zero syndrome. We use the mean-field approximation to provide differential conditions to estimate the probability of satisfying zero-syndrome conditions in Eq.~\ref{soft-syn}:
\begin{equation}
\begin{aligned}
     s_j^\dagger =\frac{1}{2}-\frac{1}{2}&\prod_{\{i:H_{j, i} = 1 \}}\Bigl(2\cdot \mathrm{sigmoid}\bigl(\frac{2 \mathbf{x}_{t, i}}{\sigma^2}\bigr) - 1\Bigr), \\ &\Pr(S_j^\dagger = 0) = 1 - s_j^\dagger,   
\end{aligned}
\end{equation}
where $x_{t, i}$ is the $i$-th position of the received codeword $x_t$, the noise level $\sigma$ is given to the decoder at the receiver's side. The soft-syndrome is zero if and only if the codeword is valid, yet it varies smoothly and continuously with the received signal $\mathbf{x}_t$. By using $e^\dagger_t$ as the time condition, we provide the consistency model with a smooth trajectory from a noisy signal to a valid codeword, resolving the instability and degeneracy issues of the hard error sum and enabling stable training. 

\subsection{Syndrome-Flow Consistency Training Dynamics}
Building upon the differential time conditions via soft-syndrome in Eq.~\ref{soft-syn}, we make it able to learn a smooth trajectory satisfying consistency conditions as shown in Figure~\ref{train}. Given a codeword $x_s\in\{-1,+1\}^n$ modulated using BPSK, and the signal received is then perturbed with an AWGN channel $y=x_s+z=x_s\cdot \tilde{z}_s$, where $\tilde{z}_s$ denotes the multiplicative noise. We follow the standard pre-processing techniques proposed by~\cite{bennatan2018deep}, the input of the neural network is a concatenated vector representing magnitude and hard syndrome $[|y|,s(y)]$ with length $2n-k$. For the forward process in AWGN channel, we build the trajectory by adding the same Gaussian noise $\epsilon\sim\mathcal{N}(0,I)$ with a different time schedule $\sqrt{\bar{\beta}_t}$, where $t\in[0,...,N]$ and $N$ denotes the pre-defined forward noising steps. Thus, during training, we sample different noisy signals $y_1:=\mathbf{x}_t,y_2:=\mathbf{x}_r$ with different noise levels $t\sim\mathcal{U}\{0,...,N\}$ and $r=\alpha t$, i.e.$\mathbf{x}_t:=\mathbf{x}_0+\sqrt{\bar{\beta}_t}\cdot\epsilon$, $\mathbf{x}_r:=\mathbf{x}_0+\sqrt{\bar{\beta}_r}\cdot\epsilon$, where $\alpha\in[0,1]$ denotes the time scaling factor. Then a consistency model $f_\theta$ predicts the clean codeword. Following Eq.~\ref{EC-CM}, we get the consistency loss for two different noisy signals $y_t,y_r$, which learns the mapping to their original clean codeword $\mathbf{x}_0$. We further add the soft-syndrome loss in Eq.~\ref{soft-syn} as the regularization term to stabilize training and fast convergence, validated in Appendix~\ref{soft_reg}:
\begin{equation}\label{total}
\begin{aligned}
\mathcal{L}_{\text{Total}}(\theta)=\mathbb{E}\Bigl[\underbrace{w(t)\left(d(f_\theta(\mathbf{x}_{t},e^\dagger_t), \mathbf{x}_0)+d\left(f_\theta(\mathbf{x}_{r},e^\dagger_r), \mathbf{x}_0\right)\right)}_{\text{Consistency Loss}}\\
+\lambda\cdot\underbrace{\left(\mathcal{L}_{\text{Soft-syn}}(f_\theta(\mathbf{x}_t,e^\dagger_t),H)+\mathcal{L}_{\text{Soft-syn}}(f_\theta(\mathbf{x}_r,e^\dagger_r),H)\right)}_{\text{Soft-syndrome Loss}}\Bigr]
\end{aligned}    
\end{equation} where $\mathcal{L}_{\text{Soft-syn}}(\mathbf{x}_t,H)=-e^\dagger_t$, $d(\cdot,\cdot)$ is a distance metric, such as Binary Cross-Entropy (BCE) in this work, and $\lambda$ is a hyperparameter that weights the syndrome regularization term. Once trained according to Algorithm~\ref{alg:training}, the learned consistency function $f_\theta$ can decode the noisy received signal $y$ in a one step: $\hat{\mathbf{x}}_0=f_\theta(y,e^\dagger_t)$, in Appendix~\ref{alg:sampling}, Algorithm~\ref{alg:sampling}.

\section{Numerical Results}
\textbf{Datasets.}  We evaluate our proposed ECCFM framework on the following set of standard error correction codes, including BCH, Polar, and LDPC codes (MacKay, CCSDS, and WRAN variants). Our evaluation considers multiple code lengths ($n$), rates ($k/n$), and Signal-to-Noise Ratios (SNRs), specifically $E_b/N_0$ values from 4 to 6 dB, to ensure a robust assessment of performance.

\textbf{Evaluation Metrics.} We evaluate decoding performance using two standard metrics following established benchmarks~\cite{choukroun2022error,choukroundenoising,parkcrossmpt}: Bit Error Rate (BER) and Frame Error Rate (FER). BER measures the fraction of individual bits that are incorrectly decoded. FER measures the fraction of entire codewords that contain one or more bit errors. Concerning the latency, we evaluate computational efficiency by reporting inference time and throughput (decoded samples per second).

\textbf{Baselines.} We numerically compared the results with multiple baselines for the decoding task, including: 1) Conventional BP-based decoders: BP~\cite{bennatan2018deep} and ARBP~\cite{nachmani2021autoregressive}. 2) Transformer-based decoders: ECCT~\cite{choukroun2022error}, and CrossMPT~\cite{parkcrossmpt}. 3) Denoising diffusion decoders: DDECC~\cite{choukroundenoising}.

\textbf{Experimental Setup.} We reproduced the results for all model-free baselines (ECCT, CrossMPT, DDECC) by implementing them with their publication-stated hyperparameters. Our primary ECCFM model utilizes a Cross-attention Transformer following CrossMPT, using $N=6$ layers and a hidden dimension of $d=128$. All the other three baselines (ECCT, CrossMPT, DDECC) apply the same network architecture to ensure fair comparison. ECCFM was trained for 1500 epochs using the Adam optimizer on a single GPU. The learning rate was managed by a cosine decay scheduler, starting at $10^{-4}$ and decreasing to $5 \times 10^{-7}$. Detailed training configurations and hyperparameter selections are provided in Appendix~\ref{setting}.

\textbf{Overall Performance}. Following established benchmarks~\cite{choukroun2022error, choukroundenoising, parkcrossmpt}, we conducted a decoding performance comparison measured in $-\ln(\text{BER})$. Our method was evaluated versus two classes of decoders: conventional BP-based algorithms (BP and ARBP) and model-free neural decoders (ECCT, CrossMPT, and DDECC). To ensure a fair comparison, all neural models were implemented with a fixed architecture ($N=6$ layers, $d=128$ hidden dimensions). Furthermore, to ensure statistical significance, each simulation was run until at least 500 error codes were observed, under a maximum of $10^8$ test instances. As shown in Table~\ref{tab:decoding_comparison}, our proposed ECCFM framework consistently achieves the best or second-best BER across the tested code families, including BCH, Polar, LDPC, CCSDS and MacKay with different code rates $(n,k)$, and showing considerable gain over POLAR codes.
\begin{table*}[!h]\huge
    \centering
    \caption{Performance comparison of various decoders across different codes and Signal-to-Noise Ratios ($E_b/N_0$). The results are reported in terms of $-\ln(\text{BER})$ (the higher, the better). All model-free methods use a fixed model architecture ($N=6$, $d=128$). Best results are shown in \textbf{bold} and the second-best results are shown in \textbf{underline}, respectively.}
    \label{tab:decoding_comparison}
    \renewcommand{\arraystretch}{1.25}
    \resizebox{\textwidth}{!}{%
        \begin{tabular}{cc ccc ccc ccc ccc ccc ccc}
        \toprule
        \multicolumn{2}{c}{\textbf{Architecture}} & \multicolumn{6}{c}{BP-based decoders} & \multicolumn{12}{c}{Model-free decoders} \\
        \cmidrule(lr){1-2} \cmidrule(lr){3-8} \cmidrule(lr){9-20}
        \multirow{2}{*}{\textbf{Code Type}} & \multirow{2}{*}{\textbf{Parameters}} & \multicolumn{3}{c}{BP} & \multicolumn{3}{c}{ARBP} & \multicolumn{3}{c}{ECCT} & \multicolumn{3}{c}{CrossMPT} & \multicolumn{3}{c}{DDECC} & \multicolumn{3}{c}{\textbf{ECCFM(Ours)}} \\
        \cmidrule(lr){3-5} \cmidrule(lr){6-8} \cmidrule(lr){9-11} \cmidrule(lr){12-14} \cmidrule(lr){15-17} \cmidrule(lr){18-20}

        & & \textbf{4} & \textbf{5} & \textbf{6} & \textbf{4} & \textbf{5} & \textbf{6} & \textbf{4} & \textbf{5} & \textbf{6} & \textbf{4} & \textbf{5} & \textbf{6} & \textbf{4} & \textbf{5} & \textbf{6} & \textbf{4} & \textbf{5} & \textbf{6}\\
        \midrule

        \multirow{2}{*}{\textbf{BCH}} 
        & (63,36) & 4.03 & 5.42 & 7.26 & 4.57 & 6.39 & 8.92 & 4.69 & 6.48 & 9.06 & 4.94 & 6.74 & 9.28 & \textbf{5.02} & \underline{6.82} & \textbf{9.88} & \underline{5.00} & \textbf{6.89} & \underline{9.76} \\
        & (63,45) & 4.36 & 5.55 & 7.26 & 4.97 & 6.90 & 9.41 & 5.47 & 7.56 & 10.51 & \textbf{5.73} & 7.98 & 10.80 & 5.68 & \textbf{8.08} & \textbf{11.22} & \underline{5.70} & \underline{8.03} & \underline{11.04}\\
        \midrule
        \multirow{5}{*}{\textbf{POLAR}} 
        & (64,32) & 4.26 & 5.38 & 6.50 & 5.57 & 7.43 & 9.82 & 6.87 & 9.21 & 12.15 & \underline{7.42} & \underline{9.94} & \underline{13.28} & 7.04 & 9.44 & 12.70 & \textbf{7.55} & \textbf{10.31} & \textbf{13.80}\\ 
        & (64,48) & 4.74 & 5.94 & 7.42 & 5.41 & 7.19 & 9.30 & 6.21 & 8.31 & 10.85 & \underline{6.36} & \underline{8.53} & \underline{11.09} & 5.93 & 8.00 & 10.44 & \textbf{6.56} & \textbf{8.78} & \textbf{11.52} \\
        & (128,64) & 4.10 & 5.11 & 6.15 & 4.84 & 6.78 & 9.30 & 5.79 & 8.45 & 11.10 & 7.45 & 9.71 & 14.31 & \underline{7.71} & \underline{11.40} & \underline{13.85} & \textbf{8.01} & \textbf{12.22} & \textbf{16.71} \\
        & (128,86) & 4.49 & 5.65 & 6.97 & 5.39 & 7.37 & 10.13 & 6.29 & 8.98 & 12.82 & 7.43 & \underline{10.80} & \underline{15.13} & \underline{7.61} & 10.50 & 13.88 & \textbf{7.78} & \textbf{11.21} & \textbf{16.05} \\
        & (128,96) & 4.61 & 5.79 & 7.08 & 5.27 & 7.44 & 10.20 & 6.30 & 9.04 & 12.40 & 7.06 & 10.25 & 13.23 & \underline{7.14} & \underline{10.31} & \underline{13.66} & \textbf{7.21} & \textbf{10.52} & \textbf{14.32} \\
        \midrule

        \multirow{3}{*}{\textbf{LDPC}} 
        & (121,60) & 4.82 & 7.21 & 10.87 & 5.22 & 8.31 & 13.07 & 5.12 & 8.21 & 12.80 & \underline{5.75} & \underline{9.42} & \underline{15.21} & 5.42 & 9.11 & 13.82 & \textbf{6.02} & \textbf{9.94} & \textbf{15.55} \\
        & (121,70) & 5.88 & 8.76 & 13.04 & 6.45 & 10.01 & 14.77 & 6.30 & 10.11 & 15.50 & \underline{7.06} & \underline{11.29} & 17.10 & 6.91 & 11.02 & \underline{17.15} & \textbf{7.35} & \textbf{12.23} & \textbf{17.60} \\
        & (121,80) & 6.66 & 9.82 & 13.98 & 7.22 & 11.03 & 15.90 & 7.27 & 11.21 & 17.02 & \underline{7.87} & \underline{12.65} & \underline{17.72} & 7.61 & 11.89 & 16.18 & \textbf{8.25} & \textbf{13.33} & \textbf{18.69} \\
        \midrule
        \textbf{MacKay} & (96,48) & 6.84 & 9.40 & 12.57 & 7.43 & 10.65 & 14.65 & 7.37 & 10.55 & 14.72 & 7.85 & 11.72 & 15.49 & \textbf{8.03} & \textbf{12.44} & \underline{15.79} & \underline{7.92} & \underline{12.25} & \textbf{16.08} \\ 
        \midrule
        \textbf{CCSDS} & (128,64) & 6.55 & 9.65 & 13.78 & 7.25 & 10.99 & 16.36 & 6.82 & 10.60 & 15.87 & 7.56 & 11.87 & 16.80 & \underline{7.77} & \underline{12.35} & \textbf{17.22} & \textbf{7.95} & \textbf{12.68} & \underline{17.01} \\ 
        \bottomrule
        \end{tabular}%
    }
\end{table*}

\begin{figure*}[h]
    \centering
    \includegraphics[width=0.95\linewidth]{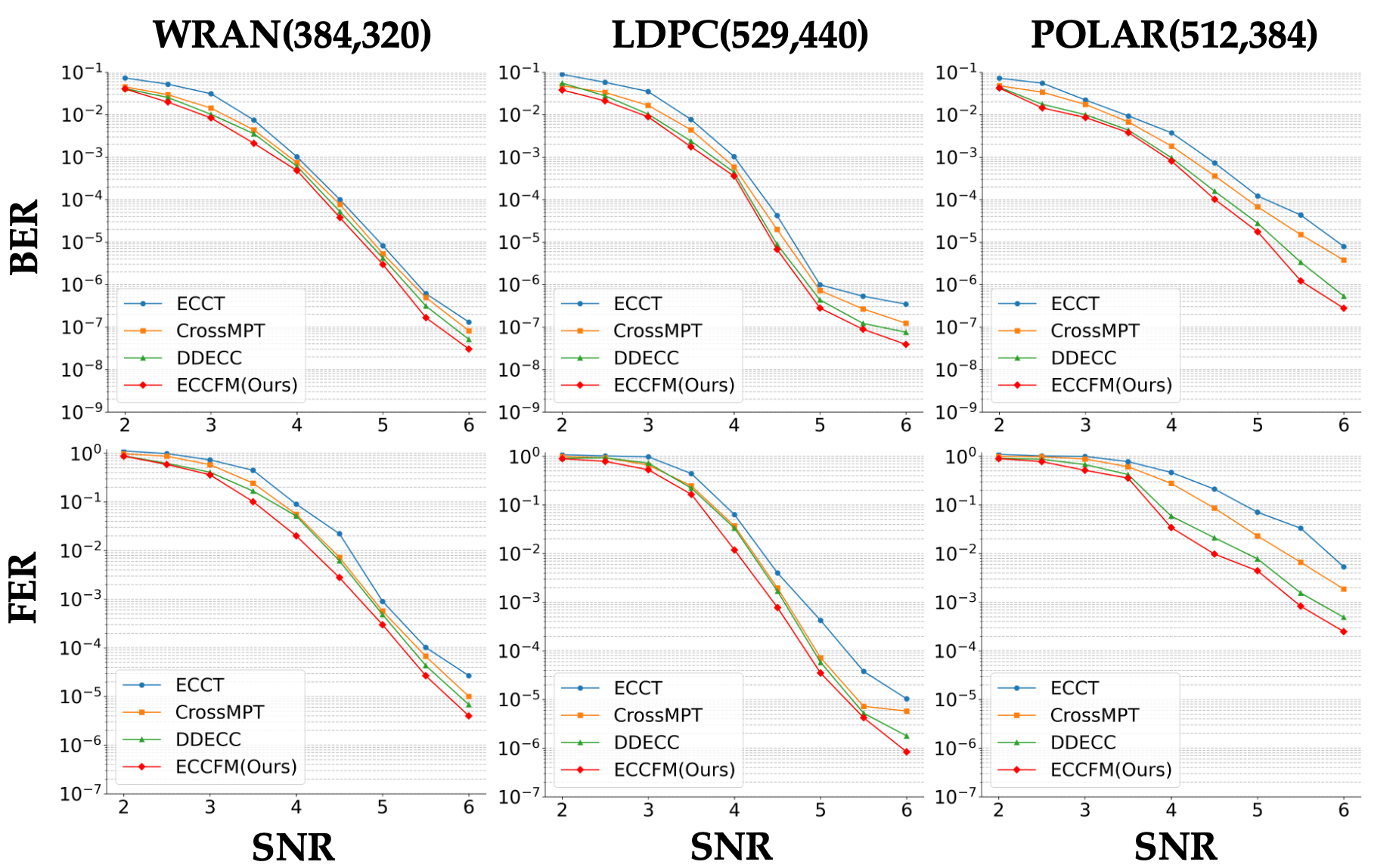}
    \caption{Performance comparison of various decoding baselines on medium-to-long block codes. The plot shows the Bit Error Rate (BER) at different Signal-to-Noise Ratios (SNRs), from 2 dB to 6 dB and divided by 0.5 dB.}
    \label{fig:longer}
\end{figure*}

\textbf{Performance on Longer Codes.} To further evaluate scalability, we conducted a evaluation on longer codes commonly used in practical communication systems: LDPC$(n=204,k=102)$, LDPC$(n=529,k=440)$, WRAN$(n=384,k=320)$, and Polar$(n=512,k=384)$. All methods were implemented with the same model architecture ($N=6,d=128$). As illustrated in Figure~\ref{fig:longer}, ECCFM improves the performance across a range of SNRs (2 dB to 6 dB). Additional results on other high-length codes are presented in Appendix~\ref{appendix_long}, which also suggest inference speed gain compared to the baselines. 

\begin{figure*}[h]
    \centering
    \includegraphics[width=0.95\linewidth]{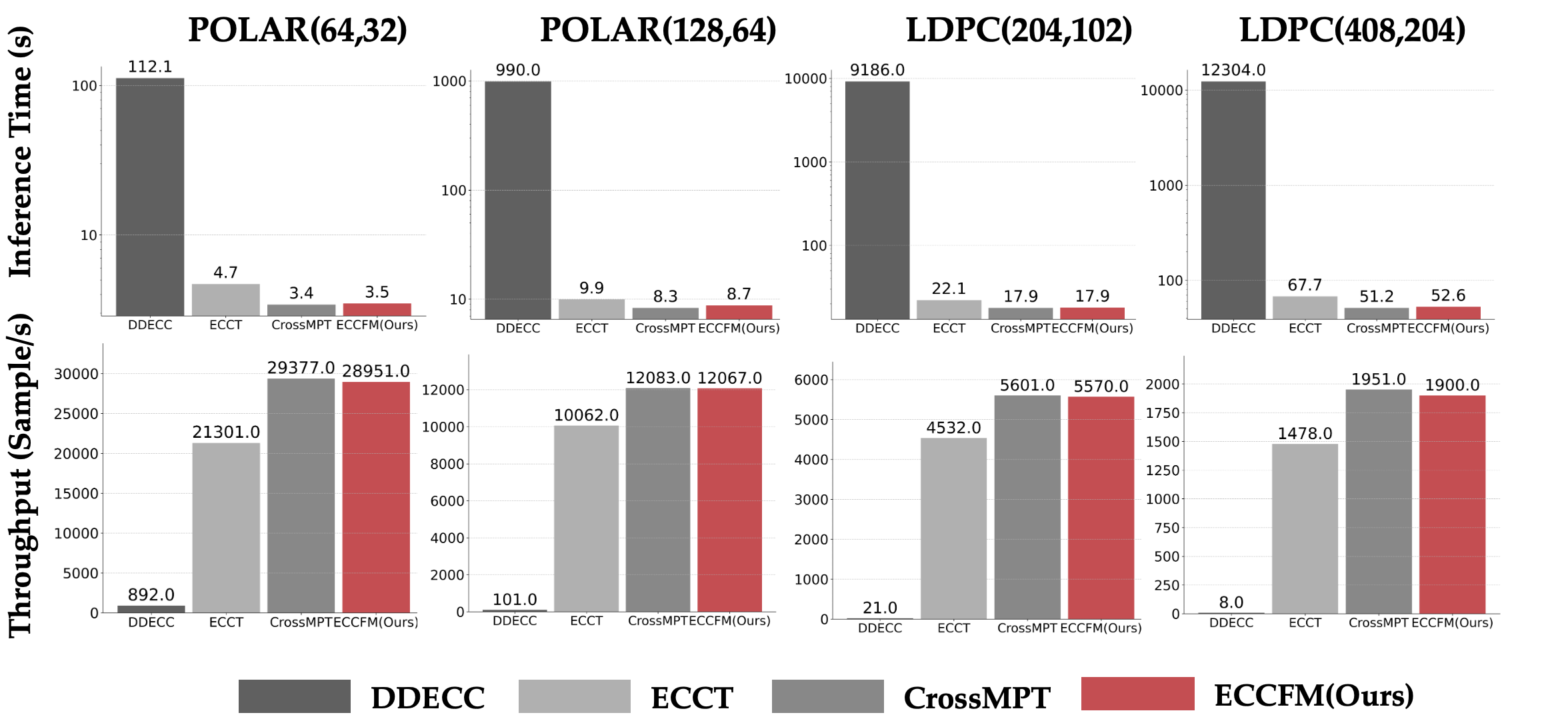}
    \caption{Comparison of Inference Time (top) and Throughput (bottom) across various decoding baselines and code types.}
    \label{fig:inference}

    \centering
    \includegraphics[width=0.95\linewidth]{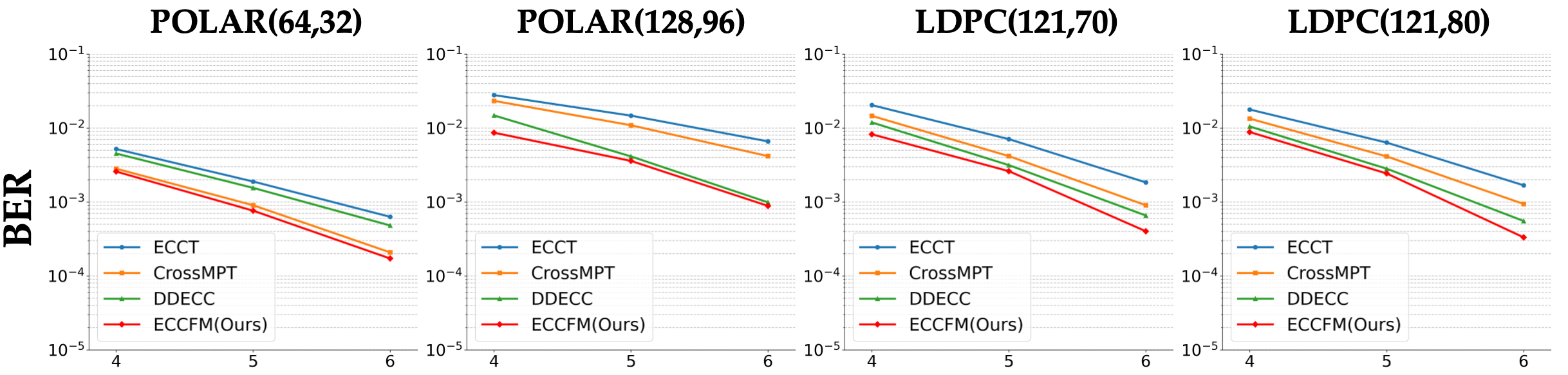}
    \caption{Performance comparison of various decoding baselines on Rayleigh Fading channels. The plot shows the Bit Error Rate (BER) at different Signal-to-Noise Ratios (SNRs).}
    \label{fig:rayleigh}
\end{figure*}

\textbf{Performance on Rayleigh Fading Channel.}
While our proposed ECCFM training dynamics and the definition of soft-syndrome are intrinsic to the code structure and independent of the underlying channel model, we validate our approach on Rayleigh fading channels to demonstrate robustness. We follow the same experimental settings described in~\cite {choukroun2022error, parkcrossmpt}. In contrast to the AWGN channel, the received signal is defined by: $y=hx+z$, where $h$ is an $ n$-dimensional i.i.d. Rayleigh distributed vector with a scale parameter $\alpha=1$ and $z\sim N(0,\sigma^2)$. In Figure~\ref{fig:rayleigh}, our method achieved competitive performance compared to baselines across different SNRs on Polar and LDPC without modifications.

\textbf{Inference Time and Throughput Comparison.} To quantify the efficiency gain, we measured inference time (total seconds to decode $10^5$ samples) and throughput (samples decoded per second). As shown in Figure~\ref{fig:inference}, ECCFM demonstrates a speed advantage over diffusion-based methods such as DDECC, achieving speedups of over \textbf{30x} for short codes and \textbf{100x} for medium-to-long codes. This disparity arises because diffusion models require several iterative denoising steps for inference, a computational cost that scales with code complexity as detailed in Appendix~\ref{appendix_inference}. Therefore, ECCFM matches the competitive performance of denoising diffusion decoders while operating at the high throughput of single-step transformer-based decoders.

\textbf{Ablation Study: Model-Agnostic Property of ECCFM.} We discussed that ECCFM is a model-agnostic training framework, i.e., its performance could be preserved over different neural network architectures. We conducted an ablation study where we decoupled our framework from the backbone. Specifically, we used the ECCT baseline architecture and trained it with our proposed ECCFM and compared this model against the original ECCT and DDECC. The results in Appendix~\ref{model-free}, Table~\ref{tab:decoding_comparison_ecct_only} show that applying the ECCFM training objective yields improvement in $-\ln(\text{BER})$ over the standard ECCT, with comparable performance versus the iterative denoising DDECC method.

\textbf{Ablation Study: Necessity of Soft-syndrome Time Condition in Consistency Training.}
We further investigated the necessity of using soft-syndrome as a time condition for building the smooth trajectory. We trained ECCFM with either a hard-syndrome time condition or a soft-syndrome time condition. In Appendix~\ref{ablation_soft}, Table~\ref{tab:eccfm_time_comparison}, our experiments revealed that using soft-syndrome as time condition is necessary for successful training of ECCFM. The continuous nature of the soft-syndrome provides a smooth trajectory to learn, whereas models conditioned on the hard syndrome failed to converge.

\section{Conclusions and Limitations}
In this work, we introduced the Error Correction Syndrome-Flow Consistency Flow Model (ECCFM), a novel training framework for obtaining high performance with low latency required for real applications. By reformulating the decoding task as a one-step consistency mapping and introducing a differential soft-syndrome condition, we handle non-smooth trajectories that previously hindered the application of consistency models to ECC. Our experiments demonstrate that ECCFM achieves comparable BER and FER across various standard codes, while offering a consistent inference speedup over denoising diffusion methods. Despite these successful results, our work has limitations to be addressed in future research. The convergence rate and training efficiency of consistency models highly rely on building a smooth trajectory, which varies in different code types. Future work can explore adaptive methods for more general decoding tasks.

\bibliographystyle{unsrt}
\bibliography{main}

@inproceedings{sohl2015deep,
  title={Deep unsupervised learning using nonequilibrium thermodynamics},
  author={Sohl-Dickstein, Jascha and Weiss, Eric and Maheswaranathan, Niru and Ganguli, Surya},
  booktitle={International conference on machine learning},
  pages={2256--2265},
  year={2015},
  organization={pmlr}
}

@article{song2020score,
  title={Score-based generative modeling through stochastic differential equations},
  author={Song, Yang and Sohl-Dickstein, Jascha and Kingma, Diederik P and Kumar, Abhishek and Ermon, Stefano and Poole, Ben},
  journal={arXiv preprint arXiv:2011.13456},
  year={2020}
}

@article{song2019generative,
  title={Generative modeling by estimating gradients of the data distribution},
  author={Song, Yang and Ermon, Stefano},
  journal={Advances in neural information processing systems},
  volume={32},
  year={2019}
}

@inproceedings{ho2020denoising,
  author    = {Jonathan Ho and Ajay Jain and Pieter Abbeel},
  title     = {Denoising Diffusion Probabilistic Models},
  booktitle = {Advances in Neural Information Processing Systems},
  volume    = {33},
  pages     = {6840--6851},
  year      = {2020},
  url       = {https://proceedings.neurips.cc/paper/2020/file/4c5bcfec8584af0d967f1ab10179ca4b-Paper.pdf}
}

@inproceedings{rombach2022high,
  author    = {Robin Rombach and Andreas Blattmann and Dominik Lorenz and Patrick Esser and Bj{\"o}rn Ommer},
  title     = {High-Resolution Image Synthesis with Latent Diffusion Models},
  booktitle = {Proceedings of the IEEE/CVF Conference on Computer Vision and Pattern Recognition (CVPR)},
  year      = {2022},
  pages     = {10684--10695},
  url       = {https://openaccess.thecvf.com/content/CVPR2022/papers/Rombach_High-Resolution_Image_Synthesis_With_Latent_Diffusion_Models_CVPR_2022_paper.pdf}
}

@article{dhariwal2021diffusion,
  title={Diffusion models beat gans on image synthesis},
  author={Dhariwal, Prafulla and Nichol, Alexander},
  journal={Advances in neural information processing systems},
  volume={34},
  pages={8780--8794},
  year={2021}
}

@article{nie2025large,
  title={Large language diffusion models},
  author={Nie, Shen and Zhu, Fengqi and You, Zebin and Zhang, Xiaolu and Ou, Jingyang and Hu, Jun and Zhou, Jun and Lin, Yankai and Wen, Ji-Rong and Li, Chongxuan},
  journal={arXiv preprint arXiv:2502.09992},
  year={2025}
}

@inproceedings{blattmann2023align,
  title={Align your latents: High-resolution video synthesis with latent diffusion models},
  author={Blattmann, Andreas and Rombach, Robin and Ling, Huan and Dockhorn, Tim and Kim, Seung Wook and Fidler, Sanja and Kreis, Karsten},
  booktitle={Proceedings of the IEEE/CVF conference on computer vision and pattern recognition},
  pages={22563--22575},
  year={2023}
}

@article{he2022latent,
  title={Latent video diffusion models for high-fidelity long video generation},
  author={He, Yingqing and Yang, Tianyu and Zhang, Yong and Shan, Ying and Chen, Qifeng},
  journal={arXiv preprint arXiv:2211.13221},
  year={2022}
}

@inproceedings{podell2024sdxl,
title={{SDXL}: Improving Latent Diffusion Models for High-Resolution Image Synthesis},
author={Dustin Podell and Zion English and Kyle Lacey and Andreas Blattmann and Tim Dockhorn and Jonas M{\"u}ller and Joe Penna and Robin Rombach},
booktitle={The Twelfth International Conference on Learning Representations},
year={2024},
url={https://openreview.net/forum?id=di52zR8xgf}
}

@article{karras2022elucidating,
  title={Elucidating the design space of diffusion-based generative models},
  author={Karras, Tero and Aittala, Miika and Aila, Timo and Laine, Samuli},
  journal={Advances in neural information processing systems},
  volume={35},
  pages={26565--26577},
  year={2022}
}

@article{sun2023difusco,
  title={Difusco: Graph-based diffusion solvers for combinatorial optimization},
  author={Sun, Zhiqing and Yang, Yiming},
  journal={Advances in neural information processing systems},
  volume={36},
  pages={3706--3731},
  year={2023}
}

@article{li2023t2t,
  title={T2t: From distribution learning in training to gradient search in testing for combinatorial optimization},
  author={Li, Yang and Guo, Jinpei and Wang, Runzhong and Yan, Junchi},
  journal={Advances in Neural Information Processing Systems},
  volume={36},
  pages={50020--50040},
  year={2023}
}

@article{lei2025boosting,
  title={Boosting Generalization in Diffusion-Based Neural Combinatorial Solver via Inference Time Adaptation},
  author={Lei, Haoyu and Zhou, Kaiwen and Li, Yinchuan and Chen, Zhitang and Farnia, Farzan},
  journal={arXiv preprint arXiv:2502.12188},
  year={2025}
}

@article{lou2023discrete,
  title={Discrete diffusion modeling by estimating the ratios of the data distribution},
  author={Lou, Aaron and Meng, Chenlin and Ermon, Stefano},
  journal={arXiv preprint arXiv:2310.16834},
  year={2023}
}

@article{song2020denoising,
  title={Denoising diffusion implicit models},
  author={Song, Jiaming and Meng, Chenlin and Ermon, Stefano},
  journal={arXiv preprint arXiv:2010.02502},
  year={2020}
}

@article{lu2022dpm,
  title={Dpm-solver: A fast ode solver for diffusion probabilistic model sampling in around 10 steps},
  author={Lu, Cheng and Zhou, Yuhao and Bao, Fan and Chen, Jianfei and Li, Chongxuan and Zhu, Jun},
  journal={Advances in neural information processing systems},
  volume={35},
  pages={5775--5787},
  year={2022}
}

@inproceedings{song2023consistency,
  title={Consistency models},
  author={Song, Yang and Dhariwal, Prafulla and Chen, Mark and Sutskever, Ilya},
  booktitle={Proceedings of the 40th International Conference on Machine Learning},
  pages={32211--32252},
  year={2023}
}

@article{geng2024consistency,
  title={Consistency models made easy},
  author={Geng, Zhengyang and Pokle, Ashwini and Luo, William and Lin, Justin and Kolter, J Zico},
  journal={arXiv preprint arXiv:2406.14548},
  year={2024}
}

@article{choukroun2022denoising,
  title={Denoising diffusion error correction codes},
  author={Choukroun, Yoni and Wolf, Lior},
  journal={arXiv preprint arXiv:2209.13533},
  year={2022}
}

@article{choukroun2022error,
  title={Error correction code transformer},
  author={Choukroun, Yoni and Wolf, Lior},
  journal={Advances in Neural Information Processing Systems},
  volume={35},
  pages={38695--38705},
  year={2022}
}

@article{richardson2002capacity,
  title={The capacity of low-density parity-check codes under message-passing decoding},
  author={Richardson, Thomas J and Urbanke, R{\"u}diger L},
  journal={IEEE Transactions on information theory},
  volume={47},
  number={2},
  pages={599--618},
  year={2002},
  publisher={IEEE}
}

@article{fossorier1999reduced,
  title={Reduced complexity iterative decoding of low-density parity check codes based on belief propagation},
  author={Fossorier, Marc PC and Mihaljevic, Miodrag and Imai, Hideki},
  journal={IEEE Transactions on communications},
  volume={47},
  number={5},
  pages={673--680},
  year={1999},
  publisher={IEEE}
}

@article{matsumine2024recent,
  title={Recent advances in deep learning for channel coding: A survey},
  author={Matsumine, Toshiki and Ochiai, Hideki},
  journal={IEEE Open Journal of the Communications Society},
  year={2024},
  publisher={IEEE}
}

@inproceedings{gruber2017deep,
  title={On deep learning-based channel decoding},
  author={Gruber, Tobias and Cammerer, Sebastian and Hoydis, Jakob and Ten Brink, Stephan},
  booktitle={2017 51st annual conference on information sciences and systems (CISS)},
  pages={1--6},
  year={2017},
  organization={IEEE}
}

@inproceedings{marinkovic2010performance,
  title={Performance evaluation of channel coding for Gbps 60-GHz OFDM-based wireless communications},
  author={Marinkovic, Miroslav and Piz, Maxim and Choi, Chang-Soon and Panic, Goran and Ehrig, Marcus and Grass, Eckhard},
  booktitle={21st Annual IEEE International Symposium on Personal, Indoor and Mobile Radio Communications},
  pages={994--998},
  year={2010},
  organization={IEEE}
}

@article{zhu2020learning,
  title={Learning to denoise and decode: A novel residual neural network decoder for polar codes},
  author={Zhu, Hongfei and Cao, Zhiwei and Zhao, Yuping and Li, Dou},
  journal={IEEE Transactions on Vehicular Technology},
  volume={69},
  number={8},
  pages={8725--8738},
  year={2020},
  publisher={IEEE}
}

@inproceedings{kim2018communication,
  title={Communication Algorithms via Deep Learning},
  author={Kim, Hyeji and Jiang, Yihan and Rana, Ranvir B and Kannan, Sreeram and Oh, Sewoong and Viswanath, Pramod},
  booktitle={International Conference on Learning Representations},
  year={2018}
}

@inproceedings{bennatan2018deep,
  title={Deep learning for decoding of linear codes-a syndrome-based approach},
  author={Bennatan, Amir and Choukroun, Yoni and Kisilev, Pavel},
  booktitle={2018 IEEE International Symposium on Information Theory (ISIT)},
  pages={1595--1599},
  year={2018},
  organization={IEEE}
}

@article{vaswani2017attention,
  title={Attention is all you need},
  author={Vaswani, Ashish and Shazeer, Noam and Parmar, Niki and Uszkoreit, Jakob and Jones, Llion and Gomez, Aidan N and Kaiser, {\L}ukasz and Polosukhin, Illia},
  journal={Advances in neural information processing systems},
  volume={30},
  year={2017}
}

@inproceedings{cammerer2017scaling,
  title={Scaling deep learning-based decoding of polar codes via partitioning},
  author={Cammerer, Sebastian and Gruber, Tobias and Hoydis, Jakob and Ten Brink, Stephan},
  booktitle={GLOBECOM 2017-2017 IEEE global communications conference},
  pages={1--6},
  year={2017},
  organization={IEEE}
}

@article{dai2021learning,
  title={Learning to decode protograph LDPC codes},
  author={Dai, Jincheng and Tan, Kailin and Si, Zhongwei and Niu, Kai and Chen, Mingzhe and Poor, H Vincent and Cui, Shuguang},
  journal={IEEE Journal on Selected Areas in Communications},
  volume={39},
  number={7},
  pages={1983--1999},
  year={2021},
  publisher={IEEE}
}

@article{kwak2022neural,
  title={Neural min-sum decoding for generalized LDPC codes},
  author={Kwak, Hee-Youl and Kim, Jae-Won and Kim, Yongjune and Kim, Sang-Hyo and No, Jong-Seon},
  journal={IEEE Communications Letters},
  volume={26},
  number={12},
  pages={2841--2845},
  year={2022},
  publisher={IEEE}
}

@article{kwak2023boosting,
  title={Boosting learning for LDPC codes to improve the error-floor performance},
  author={Kwak, Hee-Youl and Yun, Dae-Young and Kim, Yongjune and Kim, Sang-Hyo and No, Jong-Seon},
  journal={Advances in Neural Information Processing Systems},
  volume={36},
  pages={22115--22131},
  year={2023}
}

@inproceedings{lugosch2017neural,
  title={Neural offset min-sum decoding},
  author={Lugosch, Loren and Gross, Warren J},
  booktitle={2017 IEEE International Symposium on Information Theory (ISIT)},
  pages={1361--1365},
  year={2017},
  organization={IEEE}
}

@article{nachmani2019hyper,
  title={Hyper-graph-network decoders for block codes},
  author={Nachmani, Eliya and Wolf, Lior},
  journal={Advances in Neural Information Processing Systems},
  volume={32},
  year={2019}
}

@article{nachmani2021autoregressive,
  title={Autoregressive belief propagation for decoding block codes},
  author={Nachmani, Eliya and Wolf, Lior},
  journal={arXiv preprint arXiv:2103.11780},
  year={2021}
}

@inproceedings{choukroun2024foundation,
  title={A foundation model for error correction codes},
  author={Choukroun, Yoni and Wolf, Lior},
  booktitle={The Twelfth International Conference on Learning Representations},
  year={2024}
}

@inproceedings{choukroun2024learning,
  title={Learning linear block error correction codes},
  author={Choukroun, Yoni and Wolf, Lior},
  booktitle={Proceedings of the 41st International Conference on Machine Learning},
  pages={8801--8814},
  year={2024}
}

@inproceedings{parkcrossmpt,
  title={CrossMPT: Cross-attention Message-passing Transformer for Error Correcting Codes},
  author={Park, Seong-Joon and Kwak, Hee-Youl and Kim, Sang-Hyo and Kim, Yongjune and No, Jong-Seon},
  booktitle={The Thirteenth International Conference on Learning Representations},
  year={2025}
}

@inproceedings{choukroundenoising,
  title={Denoising Diffusion Error Correction Codes},
  author={Choukroun, Yoni and Wolf, Lior},
  booktitle={The Eleventh International Conference on Learning Representations},
  year={2023}
}

@article{song2023improved,
  title={Improved techniques for training consistency models},
  author={Song, Yang and Dhariwal, Prafulla},
  journal={arXiv preprint arXiv:2310.14189},
  year={2023}
}

@misc{lau2025interplaybeliefpropagationtransformer,
      title={Interplay Between Belief Propagation and Transformer: Differential-Attention Message Passing Transformer}, 
      author={Chin Wa Lau and Xiang Shi and Ziyan Zheng and Haiwen Cao and Nian Guo},
      year={2025},
      eprint={2509.15637},
      archivePrefix={arXiv},
      primaryClass={cs.IT},
      url={https://arxiv.org/abs/2509.15637}, 
}

\section{Appendix}

\subsection{Preliminary on Diffusion Generative Models}\label{cm_prelim}
\textbf{Diffusion Models.} Diffusion Models (DMs)~\cite{ho2020denoising, song2019generative, song2020score} are generative models that generate samples from a target data distribution, $p_{\text{data}}(x_0)$ by reversing a predefined forward noising process~\cite{sohl2015deep}. In the forward diffusion process, a data sample $x_0$ is gradually perturbed with Gaussian noise over a continuous time interval $t \in [0, T]$. This forward process can be mathematically described as adding noise to obtain a noisy data point $\mathbf{x}_t = \sqrt{\alpha_t} \mathbf{x}_0 + \sqrt{1 - \alpha_t} \boldsymbol{\epsilon}_t$, where $\boldsymbol{\epsilon}_t \sim \mathcal{N}(\mathbf{0}, \mathit{\boldsymbol{I}})$ is standard Gaussian noise and $\alpha_t \in [0,1]$ monotonically decreases with time step $t$ to control the noise level. Denoising Diffusion Probabilistic Models (DDPMs)~\cite{ho2020denoising} $\boldsymbol{\epsilon}_\theta:\mathcal{X} \times [T] \mapsto \mathcal{X}$ is trained to predict the noise $\boldsymbol{\epsilon}_t$ at each time step $t$, also learn the \textit{score function} of $p_t(\mathbf{x_t})$~\cite{song2019generative, song2020score}:
\begin{equation} \label{Eq: ddpm}
    \min_{\theta} \mathbb{E}_{\mathbf{x}_t, \boldsymbol{\epsilon}_t,t} \left[ \left\| \boldsymbol{\epsilon}_\theta(\mathbf{x}_t, t) - \boldsymbol{\epsilon}_t \right\|_2^2 \right] = \min_{\theta} \mathbb{E}_{\mathbf{x}_t, \boldsymbol{\epsilon}_t,t} \left[ \left\| \boldsymbol{\epsilon}_\theta(\mathbf{x}_t, t) + \sqrt{1 - \alpha_t} \underbrace{\nabla_{\mathbf{x}_t} \log p_t(\mathbf{x}_t)}_{\text{Score Function}} \right\|_2^2 \right],
\end{equation}

During inference, samples can be generated by solving the reverse-time SDE starting from $t=T$ to $t=0$. Crucially, there exists a corresponding deterministic process, the \textit{probability flow ODE} (PF-ODE), whose trajectories share the same marginal distributions ${p_t(x_t)}_{t\in[0,T]}$ as the SDE~\cite{song2020score}. The formulation of PF-ODE can be described and simplified as following~\cite{karras2022elucidating, song2023consistency}:
\begin{equation}\label{pf-ode}
    \mathrm{d}\mathbf{x}_t=-\dot\sigma(t)\sigma(t)\nabla_{\mathbf{x}_t}\log p_t(\mathbf{x}_t)\mathrm{d}t,
\end{equation}
where $\epsilon_\theta(\mathbf{x}_t, t)$ is the learned time-dependent neural network $\epsilon_\theta(\mathbf{x}_t, t)$, known as the denoiser.
, is trained to approximate this expectation: $\epsilon_\theta(x_t, t) \approx \mathbb{E}[x_0|x_t]$. By substituting this approximation and adopting the common noise schedule $\sigma(t)=t$ following~\cite{karras2022elucidating, song2023consistency}, the PF-ODE simplifies to:
\begin{equation}\label{pf-ode}
\frac{\mathrm{d}\mathbf{x}_t}{\mathrm{d}t}=-t\nabla{\mathbf{x}_t}\log p_t(\mathbf{x}_t) = \frac{\mathbf{x}_t-\epsilon_\theta(\mathbf{x}_t,t)}{t},
\end{equation}
Sample generation of diffusion models is performed by solving this PF-ODE backwards in time from $t=T$ to $t=0$, starting from a sample drawn from the prior Gaussian distribution, $\mathbf{x}_T \sim \mathcal{N}(0, \sigma(T)^2 I)$. This requires a numerical ODE solver (e.g., Euler~\cite{song2019generative, song2020score} or Heun~\cite{karras2022elucidating}) to obtain a \textit{solution trajectory} $\{\hat{\mathbf{x}}_t\}_{t\in [0,T]}$ that transforms noise into a data sample.

\textbf{Denoising Diffusion Error Correction Codes.} The application of diffusion models to error correction was pioneered by DDECC~\cite{choukroundenoising}. Its core insight is to model the transmission of a BPSK-modulated codeword $\mathbf{x}_0 \in \{-1, 1\}^n$ over an AWGN channel as the forward diffusion process. A received signal $y$ is treated as a noisy sample $\mathbf{x}_t$ at a specific timestep $t$, where the noise schedule is designed to match the channel's characteristics. This forward process is described as:
\begin{equation}\label{forward}
y := \mathbf{x}_t=\mathbf{x}_0+\sqrt{\bar{\beta_t}}\epsilon,
\end{equation}
where $\epsilon \sim \mathcal{N}(0,I)$, and the cumulative noise variance $\bar{\beta_t}=\sum_{i=1}^{t}\beta_i$ corresponds to the channel's noise level $\sigma^2$. Decoding is then performed via an iterative reverse denoising process, starting with the received signal $y:= \mathbf{x}_t$ and applying the denoising update rule for multiple steps, with a trained denoising network $\epsilon_\theta(\cdot,\cdot)$ predicting the multiplicative noise:
\begin{equation}
    \mathbf{x}_{t-1}=\mathbf{x}_t -\frac{\sqrt{\bar{\beta}_t}\beta_t}{\bar{\beta}_t+\beta_t}(\mathbf{x}_t-\text{sign}(\mathbf{x}_t)\epsilon_\theta(\mathbf{x}_t,t)),
\end{equation}
A key innovation in DDECC is its conditioning mechanism. In the ECC domain, the number of parity check errors (syndrome sum), $e_t = \sum_i s(y)_i$, serves as a direct measure of the noisy level in a received signal $y$. Therefore, DDECC conditions its denoising network $\epsilon_\theta$ on the sum of syndrome error $e_t$ instead of a timestep $t$, making the diffusion models adapted to the structure of the error correction problem. The denoising network is trained to learn the hard prediction of the multiplicative noise with a Binary Cross-Entropy (BCE) loss:
\begin{equation}\label{ddecc_loss}
    \mathcal{L}(\theta)=-\mathbb{E}_{e_t,\mathbf{x}_0,\epsilon}\log(\epsilon_\theta(\mathbf{x}_0+\sqrt{\bar{\beta}_{t}}\epsilon,e_t),\tilde{\epsilon}_b),
\end{equation}
where $\tilde{\epsilon}_b=\text{bin}(\mathbf{x}_0(\mathbf{x}_0+\sqrt{\bar{\beta}_t}\epsilon))$ denotes the target binary multiplicative noise.

\textbf{Consistency Models (CMs).} A major concern of DMs is the slow sampling process, which requires sequential calculation of the denoiser $\epsilon_\theta$. Consistency Models (CMs)~\cite{song2023consistency} were introduced to overcome this by enabling fast, 1-step generation. The core principle is the self-consistency property: any two points $(\mathbf{x}_t, t)$ and $(\mathbf{x}_{r}, r)$ on the same PF-ODE trajectory should map to the same origin point, $\mathbf{x}_0$. CMs build upon Eq.~\ref{pf-ode} and learn a function $f_{\theta}(\mathbf{x}_t, t)$ that directly estimates the \textit{trajectory} from noisy data to clean data with a single step:
\begin{equation}\label{cm}
    f_\theta(\mathbf{x}_t,t)=\mathbf{x}_0,
\end{equation}
The training objective of CMs is to enforce the consistency property across a discrete set of time steps. The continuous time interval $[0, T]$ is discretized into $N-1$ sub-intervals, defined by timesteps $1=t_1 < \dots < t_N=T$. The model is then trained to minimize the following loss, which enforces that the model's output is consistent for adjacent points on the same ODE trajectory:
\begin{equation}
\mathop{\arg\min}_{\theta}\,\mathbb{E}[w(t_i)d(f_\theta(\mathbf{x}_{t_{i+1}},t_{i+1}),f_{\theta-}(\tilde{\mathbf{x}}_{t_{i}},t_{i}))],
\end{equation}
Here, $f_\theta$ is the network being trained, while $f_{\theta-}$ is an exponential moving average (EMA) of $f_\theta$'s past samples. The term $\tilde{\mathbf{x}}_{t_{i}}=\mathbf{x}_{t_{i+1}}-(t_i-t_{i+1})t_{i+1}\nabla_{\mathbf{x}_{t_{i+1}}}\log p_{t_{i+1}}(\mathbf{x}_{t_{i+1}})$ is obtained by taking a single ODE solver step backwards from $\mathbf{x}_{t_{i+1}}$ using the score function. This training process can be performed in two ways: by distilling knowledge from a pre-trained diffusion model, known as Consistency Distillation (CD), or by training from scratch, known as Consistency Training (CT). However, training CMs is difficult and resource-intensive. It requires a carefully designed curriculum for the number of discretization steps $N$ to ensure stabilized training. The follow-up works improved the vanilla CMs, such as iCT~\cite{song2023improved}, which proposed enhanced metrics and schedulers, and ECT~\cite{geng2024consistency}, which uses "pre-training diffusion $+$ consistency tuning" to stabilize learning.

\subsection{Consistency Sampling Algorithm}
Once we obtain the well-trained consistency model $f_\theta$ following Algorithm~\ref{alg:training}, we can simply apply the one-step sampling to estimate the codeword $\hat{\mathbf{x}}_0$, given any received signals $y$.

\begin{algorithm}[h]
\caption{Error Correction Consistency One-step Sampling}
\label{alg:sampling}
\begin{algorithmic}
\Require{Consistency Model $f_{\theta}$, parity-check matrix $\mathbf{H}$.}
\For{Test batch noisy signals $y$}
    \State $e^\dagger_t=\mathcal{L}_{\text{Soft-syn}}(y,H)$
    \Comment{Calculate soft-syndrome}
    \State $\hat{\mathbf{x}}_0=f_\theta(y,e^\dagger_t)$
    \Comment{Estimate clean codeword with one-step}
\EndFor
\end{algorithmic}
\end{algorithm}

\subsection{Proof of Proposition~\ref{prop}}
\begin{proof}
Let $\hat{\mathbf{x}}_t = f_\theta(\mathbf{x}_{t},t)$ and $\hat{\mathbf{x}}_r = f_\theta(\mathbf{x}_{r},r)$ denote the predicted outcome by consistency model at timesteps $t$ and $r$, respectively. Let $\mathcal{L}_{\text{Standard-CM}}$ be defined by the Total Variation distance, $\text{TV}(\cdot,\cdot)$, and $\mathcal{L}_{\text{EC-CM}}$ be defined by the Binary Cross Entropy, $\text{BCE}(\cdot,\cdot)$. Let $\mathbf{x}_0$ denote the ground truth codeword.

First, by the Triangle Inequality of the Total Variation (TV) distance, we have:
\begin{equation}
\mathcal{L}_{\text{Standard-CM}}=\text{TV}(\hat{\mathbf{x}}_t, \hat{\mathbf{x}}_r) \le \text{TV}(\hat{\mathbf{x}}_t, \mathbf{x}_0) + \text{TV}(\hat{\mathbf{x}}_r, \mathbf{x}_0).
\end{equation}
Squaring both sides of the inequality:
\begin{equation}
\begin{aligned}
\text{TV}^2(\hat{\mathbf{x}}_t, \hat{\mathbf{x}}_r) &\le \left(\text{TV}(\hat{\mathbf{x}}_t, \mathbf{x}_0) + \text{TV}(\hat{\mathbf{x}}_r, \mathbf{x}_0)\right)^2\\
&\le 2 \text{TV}^2(\hat{\mathbf{x}}_t, \mathbf{x}_0) + 2 \text{TV}^2(\hat{\mathbf{x}}_r, \mathbf{x}_0),
\end{aligned}
\label{eq:tv_bound}
\end{equation}
By Pinsker's Inequality, which states that for any two probability distributions $P$ and $Q$, $D_{\text{KL}}(P \| Q) \ge 2 \text{TV}^2(P, Q)$,
\begin{equation}
\text{TV}^2(\hat{\mathbf{x}}_t, \hat{\mathbf{x}}_r) \le D_{\text{KL}}(\mathbf{x}_0 \| \hat{\mathbf{x}}_t)+D_{\text{KL}}(\mathbf{x}_0 \| \hat{\mathbf{x}}_r),
\end{equation}
Recall that the BCE loss is defined as $\text{BCE}(P, Q) = H(Q) + D_{\text{KL}}(Q \| P)$, where $H(Q)$ is the entropy of $Q$. Since entropy is non-negative ($H(Q) \ge 0$), we have:
\begin{equation}
D_{\text{KL}}(\mathbf{x}_0 \| \hat{\mathbf{x}}) \le \text{BCE}(\hat{\mathbf{x}}, \mathbf{x}_0),
\end{equation}
Substituting them into Eq.~\ref{eq:tv_bound}, we show that:
\begin{equation}
    \text{TV}^2(\hat{\mathbf{x}}_t, \hat{\mathbf{x}}_r) \le \text{BCE}(\hat{\mathbf{x}}_t, \mathbf{x}_0) + \text{BCE}(\hat{\mathbf{x}}_r, \mathbf{x}_0)=\mathcal{L}_{\text{EC-CM}}.
\end{equation}
\end{proof}

\subsection{Detailed Experimental Setting}\label{setting}
We provide the specific design choices of ECCFM and the listed training hyperparameters. Following the design in~\cite{choukroun2022error, choukroundenoising, parkcrossmpt}, we generate the training set by corrupting the all-zero codeword $\mathbf{x}_0$ with AWGN noise $z$, following the diffusion process $y = \mathbf{x}_0 + \sqrt{\bar{\beta}_N} \cdot \epsilon$, where $\epsilon \sim \mathcal{N}(0,I)$. The diffusion process was configured with a total of $N = n-k+5$ steps. We employed a linear variance schedule with $\beta_i = 0.01$ for short codes and a more fine-grained $\beta_i = 0.0025$ for medium-to-long codes. Each epoch consisted of 1,000 steps with a minibatch size of 128. To ensure a fair comparison, all neural models were implemented with a fixed architecture ($N=6$ layers, $d=128$ hidden dimensions). Each test task was run until at least 500 error codes were observed, under a maximum of $10^7$ test instances.

\begin{table}[htbp]
    \centering
    \caption{Training hyperparameters and design choices.} 
    \label{tab:training_choices} 
    \resizebox{0.99\textwidth}{!}{%
    \begin{tabular}{ll}
        \toprule
        \textbf{Parameters} & \textbf{Design Choice} \\
        \midrule
        Consistency Loss & $d(\cdot,\cdot) = \text{Binary Cross Entropy}(\hat{\mathbf{x}}_0, \mathbf{x}_0)$ \\
        Syndrome Weight & $\lambda=0.01$ \\
        Training Epoch & $1500$ \\
        Mini-batch & $1000$ \\
        Training Batchsize & $128$ \\
        Test Numbers & At least 500 error cases and at most $10^7$ total numbers \\
        Test Batchsize & $2048$ for short codes, $256$ for medium-to-long codes \\
        Weighting Function        & $w(t) = 1$ \\
        Total Diffusion Steps & $N=n-k+5$\\
        Forward Schedule & $\beta_i=0.01$ for short codes, $\beta_i=0.0025$ for medium-to-long codes \\
        Time Step & $t \sim \mathcal{U}\{1, 2, \dots, N\}$\\
        Scaling Factor            & $\alpha = 0.8$ \\
        Initial Learning Rate     & $\eta = 1e^{-4}$ \\
        Learning Rate Schedule    & Cosine Decay \\
        Decay Rate & $\eta' = 5e^{-7}$ \\
        Exponential Moving Average Ratio & EMA$=0.999$ \\
        \bottomrule
    \end{tabular}}
\end{table}

\subsection{Additional experimental results}\label{addition}

\subsubsection{Additional Results on Multi-step Sampling}
We investigated the multi-step sampling technique from~\cite{song2023consistency} to explore the quality-cost trade-off. Using a 2-step strategy with a re-noising ratio of $\alpha=0.2N$, we observed that performance slightly deteriorated (higher BER) compared to one-step decoding in Table~\ref{tab:multi-step}. 

We interpret this result from the nature of ECC task. Unlike image generation, error correction aims to recover the exact original data from different noise levels, and the multiple decoding steps for a one-step consistency decoder risk introducing additional error rather than refining the result, potentially leading to error accumulation.
\begin{table}[h!]
    \centering
    \caption{Performance comparison of ECCFM(1-step) and ECCFM(2-step)}
    \label{tab:multi-step}
    \renewcommand{\arraystretch}{1.25}
    \begin{tabular}{l ccc ccc}
    \toprule
    \multirow{2}{*}{\textbf{Code Type}} & \multicolumn{3}{c}{\textbf{ECCFM (1-step)}} & \multicolumn{3}{c}{\textbf{ECCFM (2-step)}} \\
    \cmidrule(lr){2-4} \cmidrule(lr){5-7} 
    $\boldsymbol{E_b/N_0}$ & \textbf{4} & \textbf{5} & \textbf{6} & \textbf{4} & \textbf{5} & \textbf{6} \\
    \midrule
    POLAR(64,32)   & \textbf{7.55} & \textbf{10.31} & \textbf{13.80} & 7.21 & 9.97 & 13.22 \\
    POLAR(128,64)  & \textbf{8.01} & \textbf{12.22} & \textbf{16.71} & 7.67 & 11.89 & 16.02 \\
    POLAR(128,96)  & \textbf{7.21} & \textbf{10.52} & \textbf{14.32} & 6.97 & 10.16 & 13.88 \\
    \midrule
    LDPC(121,70)   & \textbf{7.35} & \textbf{12.23} & \textbf{17.60} & 7.02 & 11.17 & 15.93 \\
    LDPC(121,80)   & \textbf{8.25} & \textbf{13.33} & \textbf{18.69} & 7.95 & 13.14 & 17.78 \\
    \bottomrule
    \end{tabular}
\end{table}

\subsubsection{Soft-syndrome Under Low SNR}\label{abl_snr}
\begin{figure}[htbp]
    \centering
    \subfloat[Syndrome Change on LDPC(64,48)]{\label{fig:a}\includegraphics[width=0.48\textwidth]{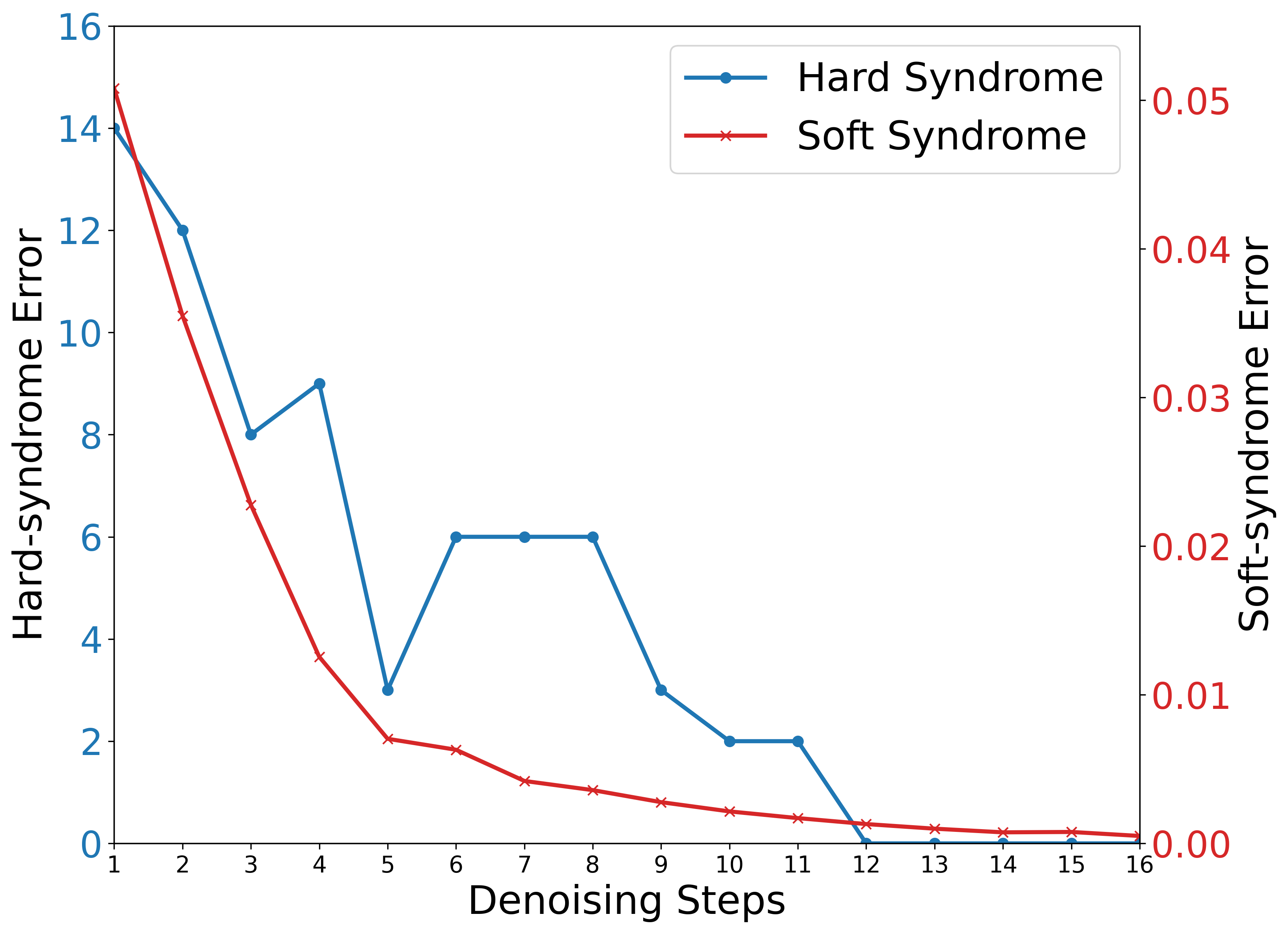}}
    \subfloat[Syndrome Change on BCH(63,36)]{\label{fig:b}\includegraphics[width=0.48\textwidth]{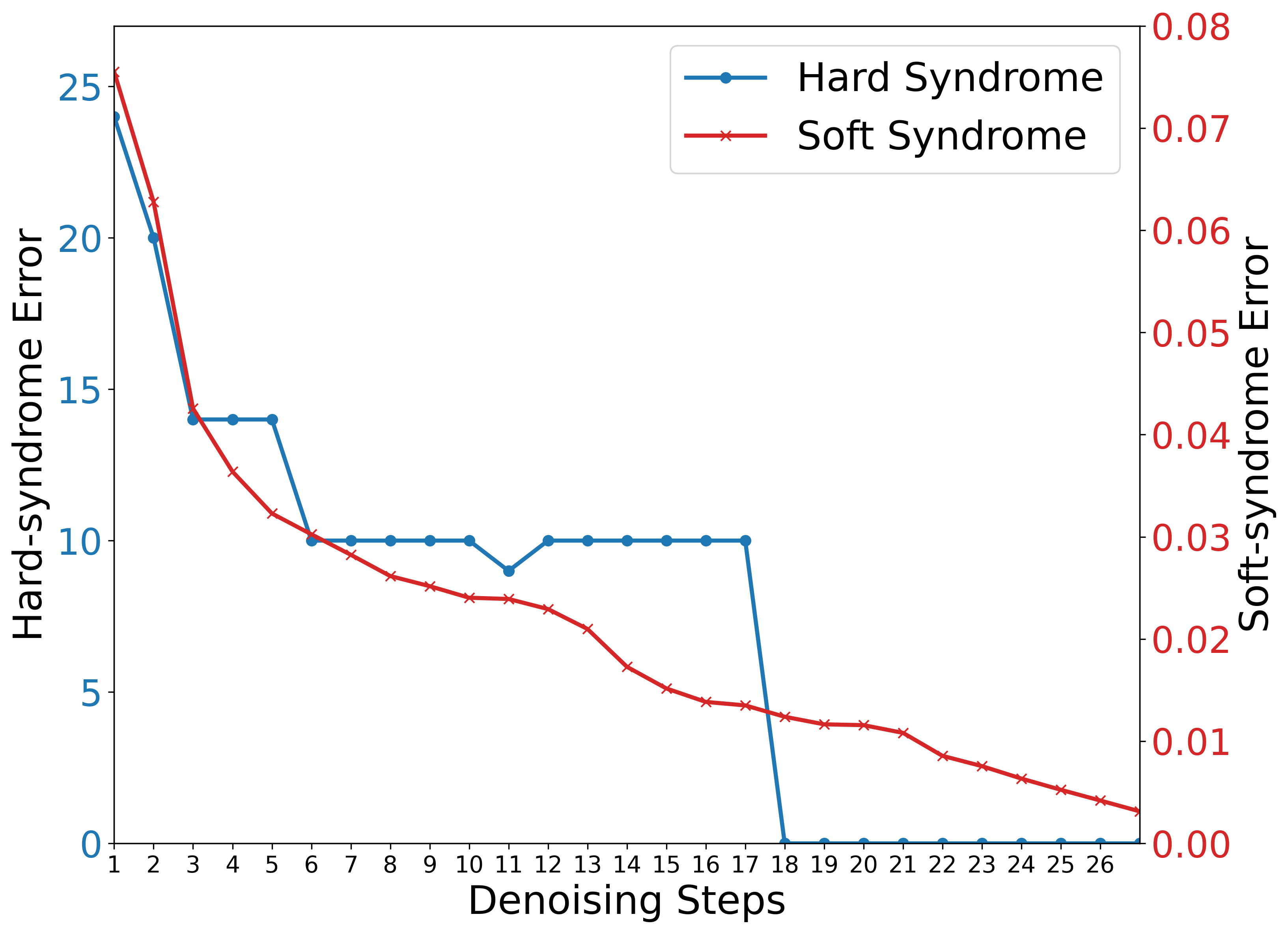}}

    \caption{Syndrome Change (Hard syndrome and soft syndrome) during iterative denoising on LDPC and BCH under low SNRs (2 dB).}
    \label{fig:syn_images}
\end{figure}

The soft-syndrome time condition in the diffusion denoising phase does not require any theoretical assumption on SNRs. To validate the efficacy of soft-syndrome condition under low SNR, we visualize the syndrome change for both hard-syndrome and soft-syndrome for POLAR(64,48) and BCH(63,36) codes under 2 dB in Figure~\ref{fig:syn_images}. The results show that soft-syndrome maintains consistent performance in building a smooth trajectory.

\subsubsection{Quantifying the Training Costs}
In the main part, we apply the cross-attention transformer proposed by~\cite{parkcrossmpt} as the backbone architecture for ECCFM. The total number of parameters marginally exceeds that of the standard CrossMPT backbone due to a 2-layer time embedding for the soft-syndrome time condition. We present a comparison of the training costs in Table~\ref{tab:training}.
\begin{table}
\caption{Quantifying the training costs of ECCFM and CrossMPT.}
\renewcommand{\arraystretch}{1.25}
\resizebox{\textwidth}{!}{ 
    \begin{tabular}{cc cc cc cc}
    \toprule
    \multirow{2}{*}{\textbf{Code Type}} & \multirow{2}{*}{\textbf{Parameter}} & \multicolumn{2}{c}{\textbf{FLOPs}} & \multicolumn{2}{c}{\textbf{Model Parameters}} & \multicolumn{2}{c}{\textbf{Training (epoch)}} \\
    \cmidrule(lr){3-4} \cmidrule(lr){5-6} \cmidrule(lr){7-8}
     & & \textbf{CrossMPT} & \textbf{ECCFM} & \textbf{CrossMPT} & \textbf{ECCFM} &\textbf{CrossMPT} & \textbf{ECCFM} \\
    \midrule
    \textbf{BCH} & (61,36) & 112.2 M & 112.6 M & 1.20 M & 1.42 M & 41 s & 71 s \\
    \midrule
    \multirow{3}{*}{\textbf{POLAR}} & (64,32) & 120.4 M & 120.8 M & 1.20 M & 1.42 M & 41 s & 75 s \\
    & (128,64) & 253.4 M & 253.8 M & 1.24 M & 1.45 M & 43 s & 78 s \\
    & (128,96) & 202.8 M & 203.2 M & 1.23 M & 1.44 M & 47 s & 82 s \\
    \midrule
    \multirow{2}{*}{\textbf{LDPC}} & (121,70) & 229.6 M & 230.0 M & 1.23 M & 1.44 M & 43 s & 67 s \\ & (121,80) & 212.4 M & 212.9 M & 1.23 M & 1.44 M & 45 s & 69 s \\
    \midrule
    \textbf{WRAN} & (384,320) & 608.2 M & 608.6 M & 1.42 M & 1.63 M & 74 s & 83 s \\
    \bottomrule
    \end{tabular}
}
\label{tab:training}
\end{table}

\subsubsection{Ablation Study: Model-Agnostic Property of ECCFM.}\label{model-free}
We discussed that ECCFM is a model-agnostic training framework, i.e., its performance could be preserved over different neural network architectures. We conducted an ablation study where we decoupled our framework from the backbone. Specifically, we used the ECCT baseline architecture and trained it with our proposed ECCFM. We then compared this model directly against the original ECCT and DDECC. The results in Table~\ref{tab:decoding_comparison_ecct_only} show that applying the ECCFM training objective yields improvement in $-\ln(\text{BER})$ over the standard ECCT, with comparable performance versus the iterative denoising DDECC method.
\begin{table}[h]
    \centering
    \caption{Performance comparison of ECCFM versus the standard ECCT and DDECC, using an identical ECCT backbone on POLAR and LDPC codes.}
    \label{tab:decoding_comparison_ecct_only}
    \renewcommand{\arraystretch}{1.125}
    \resizebox{0.9\textwidth}{!}{
    \begin{tabular}{cc ccc ccc ccc} 
        \toprule
        \multicolumn{2}{c}{\textbf{Architecture}} & \multicolumn{9}{c}{\textbf{ECCT Backbone}} \\
        \cmidrule(lr){1-2} \cmidrule(lr){3-11}
        \multirow{2}{*}{\textbf{Code}} & \multirow{2}{*}{\textbf{Parameters}} & \multicolumn{3}{c}{ECCT} & \multicolumn{3}{c}{DDECC} & \multicolumn{3}{c}{\textbf{ECCFM(ECCT)}} \\
        \cmidrule(lr){3-5} \cmidrule(lr){6-8} \cmidrule(lr){9-11}

        & & \textbf{4} & \textbf{5} & \textbf{6} & \textbf{4} & \textbf{5} & \textbf{6} & \textbf{4} & \textbf{5} & \textbf{6} \\
        \midrule

        \multirow{5}{*}{\rotatebox{90}{\textbf{POLAR}}} 
        & (64,32)  & 6.87 & 9.21 & 12.15 & \underline{7.04} & \underline{9.44} & \underline{12.70} & \textbf{7.12} & \textbf{9.77} & \textbf{12.71} \\ 
        & (64,48)  & \underline{6.21} & \underline{8.31} & \underline{10.85} & 5.93 & 8.00 & 10.44 & \textbf{6.38} & \textbf{8.55} & \textbf{11.23} \\ 
        & (128,64) & 5.79 & 8.45 & 11.10 & \textbf{7.71} & \textbf{11.40} & \underline{13.85} & \underline{7.32} & \underline{11.03} & \textbf{14.87}\\
        & (128,86) & 6.29 & 8.98 & 12.82 & \textbf{7.61} & \textbf{10.50} & \underline{13.88} & \underline{7.18} & \underline{10.17} & \textbf{15.02}\\
        & (128,96) & 6.30 & 9.04 & 12.40 & \textbf{7.14} & \textbf{10.31} & \underline{13.66} & \underline{6.86} & \underline{9.94} & \textbf{13.83} \\
        \midrule

        \multirow{3}{*}{\rotatebox{90}{\textbf{LDPC}}} 
        & (121,60) & 5.12 & 8.21 & 12.80 & \underline{5.42} & \textbf{9.11} & \underline{13.82} & \textbf{5.55} & \underline{8.86} & \textbf{13.97} \\
        & (121,70) & 6.30 & 10.11 & 15.50 & \textbf{6.91} & \underline{11.02} & \textbf{17.15} & \underline{6.87} & \textbf{11.21} & \underline{16.13} \\
        & (121,80) & 7.27 & 11.21 & \underline{17.02} & \underline{7.61} & \underline{11.89} & 16.18 & \textbf{7.80} & \textbf{12.03} & \textbf{17.95} \\
        \bottomrule
    \end{tabular}}
    \vspace{-0.5cm}
\end{table}

\subsubsection{Ablation Study: Soft-syndrome Loss Term in Total Loss}\label{soft_reg}
The soft-syndrome loss term in Eq.~\ref{total} serves primarily as a regularization term, designed to stabilize training and accelerate convergence. By explicitly incorporating this loss, we guide the optimization toward a feasible solution space during the early stages of training. To empirically validate this, we conducted an ablation study comparing the model's performance with and without the soft-syndrome term. As shown in Table~\ref{tab:comparison}, including this loss results in faster convergence of the primary consistency objective.
\begin{table}[h]
\centering
\caption{Comparison of training time to achieve the target consistency loss with or without soft-syndrome loss term}
\label{tab:comparison}
\begin{tabular}{llcc}
\toprule
\textbf{Code Type} & \textbf{Loss function} & \textbf{Target consistency loss} & \textbf{Epoch} \\
\midrule
\multirow{2}{*}{POLAR(64,32)} & \textbf{w/ Soft-syn Loss} & $1.05 \times 10^{-4}$ & \textbf{721} \\
                                & w/o Soft-syn Loss & $1.05 \times 10^{-4}$ & 1323 \\
\midrule
\multirow{2}{*}{POLAR(128,64)} & \textbf{w/ Soft-syn Loss} & $2.06 \times 10^{-2}$ & \textbf{866} \\
                                & w/o Soft-syn Loss & $2.06 \times 10^{-2}$ & 1474 \\
\midrule
\multirow{2}{*}{LDPC(121,80)} & \textbf{w/ Soft-syn Loss} & $6.50 \times 10^{-3}$ & \textbf{827} \\
                                & w/o Soft-syn Loss & $6.50 \times 10^{-3}$ & 1451 \\
\bottomrule
\end{tabular}
\end{table}

\subsubsection{Additional Results of Soft-syndrome Time Condition in Consistency Training}\label{ablation_soft}
As stated in the main part, we trained ECCFM with either a hard-syndrome time condition or a soft-syndrome time condition and validated the necessity of using soft-syndrome as time condition for successful training of ECCFM in Table~\ref{tab:eccfm_time_comparison}.
\begin{table}[h!]
    \centering
    \caption{Performance comparison of ECCFM(Hard-syndrome) and ECCFM(Soft-syndrome)}
    \label{tab:eccfm_time_comparison}
    \renewcommand{\arraystretch}{1.25}
    \begin{tabular}{l ccc ccc}
    \toprule
    \multirow{2}{*}{\textbf{Code Type}} & \multicolumn{3}{c}{\textbf{Hard-syndrome}} & \multicolumn{3}{c}{\textbf{Soft-syndrome}} \\
    \cmidrule(lr){2-4} \cmidrule(lr){5-7} 
    $\boldsymbol{E_b/N_0}$ & \textbf{4} & \textbf{5} & \textbf{6} & \textbf{4} & \textbf{5} & \textbf{6} \\
    \midrule
    POLAR(64,32)   & 4.36 & 5.78 & 9.81 & \textbf{7.55}& \textbf{10.31}& \textbf{13.80} \\
    POLAR(128,64)  & 4.87 & 7.92 & 10.10 & \textbf{8.01}& \textbf{12.22}& \textbf{16.71} \\
    POLAR(128,96)  & 4.11 & 6.97 & 9.22 & \textbf{7.21}& \textbf{10.52}& \textbf{14.32} \\
    \midrule
    LDPC(121,70)   & 4.32 & 8.31 & 11.41& \textbf{7.35} & \textbf{12.23} & \textbf{17.60} \\
    LDPC(121,80)   & 5.09 & 8.95 & 11.90 & \textbf{8.25}& \textbf{13.33}& \textbf{18.69} \\
    \bottomrule
    \end{tabular}
\end{table}

\subsubsection{Applying Vanilla Consistency Training in ECC}\label{vanilla-train} To clarify the contribution of our proposed consistency training framework, we conducted an ablation study comparing ECCFM against a baseline utilizing the standard consistency model (Vanilla-CM) objective. In this experiment, we retained the forward sampling and time-conditioning mechanisms from DDECC but replaced the training objective with the standard consistency loss following~\cite{song2023consistency}. As shown in Table~\ref{tab:eccfm_loss_comparison}, directly applying the Vanilla-CM to the ECC domain results in a significant performance degradation, and highlights a fundamental mismatch between the standard consistency objective and the discrete, non-differentiable nature of the syndrome condition in ECC domain.

\begin{table}[h!]
    \centering
    \caption{Performance comparison of Vanilla-CM and ECCFM}
    \label{tab:eccfm_loss_comparison}
    \renewcommand{\arraystretch}{1.25}
    \begin{tabular}{l ccc ccc}
    \toprule
    \multirow{2}{*}{\textbf{Code Type}} & \multicolumn{3}{c}{\textbf{Vanilla-CM}} & \multicolumn{3}{c}{\textbf{ECCFM}} \\
    \cmidrule(lr){2-4} \cmidrule(lr){5-7} 
    $\boldsymbol{E_b/N_0}$ & \textbf{4} & \textbf{5} & \textbf{6} & \textbf{4} & \textbf{5} & \textbf{6} \\
    \midrule
    POLAR(64,32)   & 4.67& 6.21& 8.32&\textbf{7.55}& \textbf{10.31}& \textbf{13.80} \\
    POLAR(128,64)  & 4.98& 6.45& 8.78&\textbf{8.01}& \textbf{12.22}& \textbf{16.71} \\
    POLAR(128,96)  & 4.33& 5.86& 7.91&\textbf{7.21}& \textbf{10.52}& \textbf{14.32} \\
    \midrule
    LDPC(121,70)   & 4.21 & 6.13 & 8.45 & \textbf{7.35} & \textbf{12.23} & \textbf{17.60} \\
    LDPC(121,80)   & 4.87& 6.84& 9.11&\textbf{8.25}& \textbf{13.33}& \textbf{18.69} \\
    \bottomrule
    \end{tabular}
\end{table}

\subsubsection{Additional Results on Long Codes} \label{appendix_long}
\begin{figure}[htbp]
    \centering
    \subfloat[BER on LDPC(204,102)]{\label{fig:a}\includegraphics[width=0.48\textwidth]{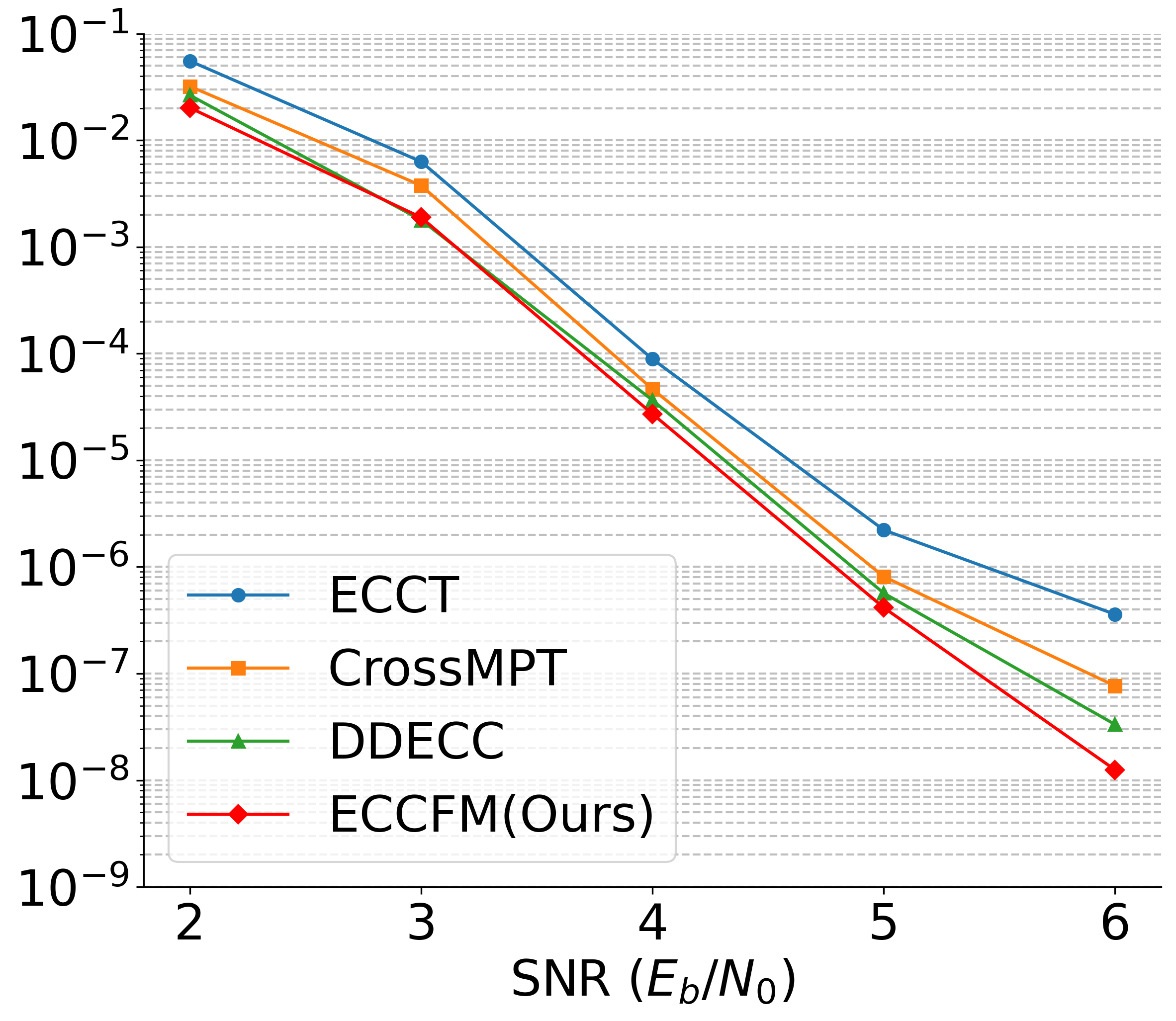}}
    \subfloat[BER on LDPC(408,204)]{\label{fig:b}\includegraphics[width=0.48\textwidth]{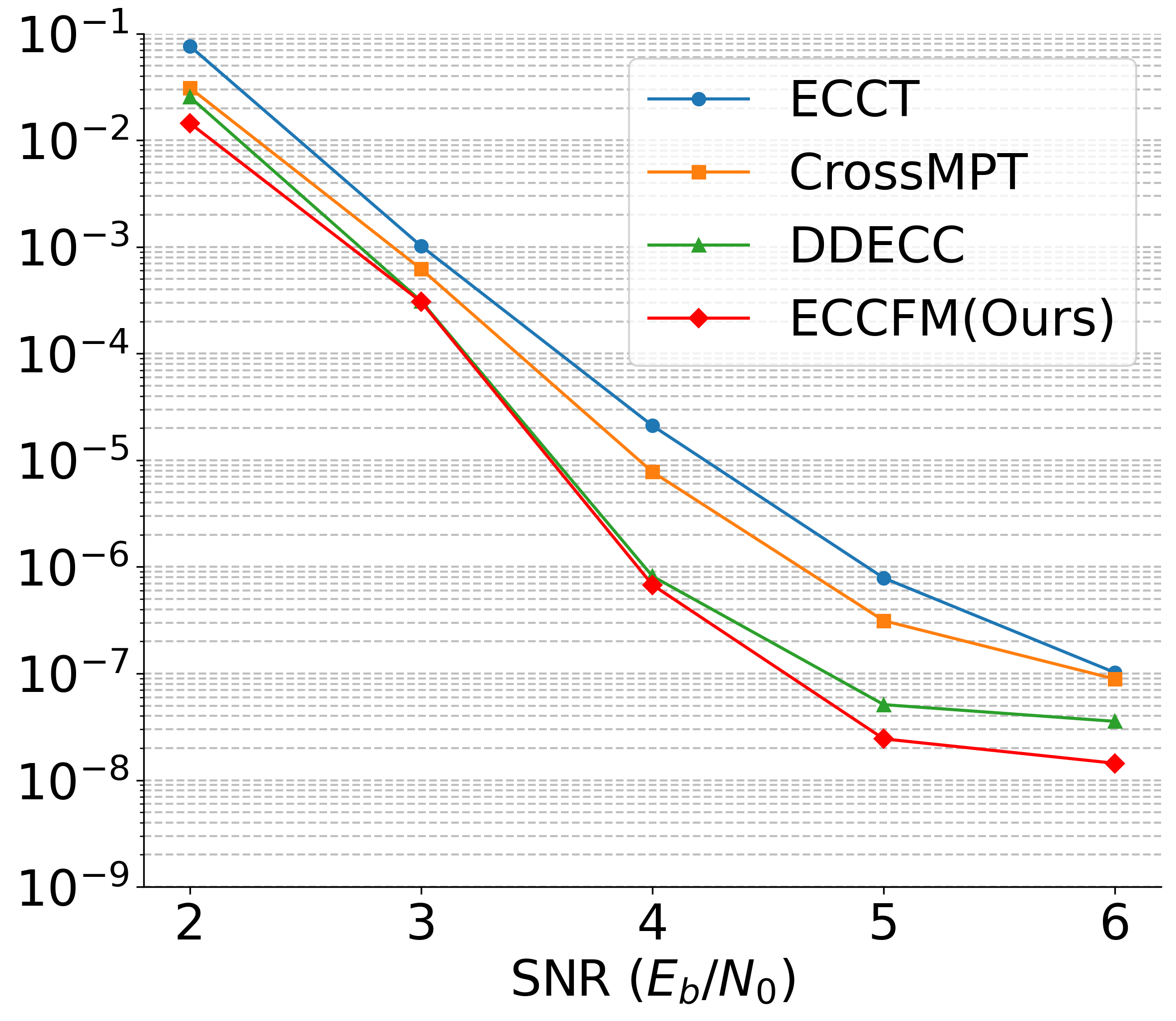}}
    \\[0.5cm]
    \subfloat[FER on LDPC(204,102)]{\label{fig:a}\includegraphics[width=0.48\textwidth]{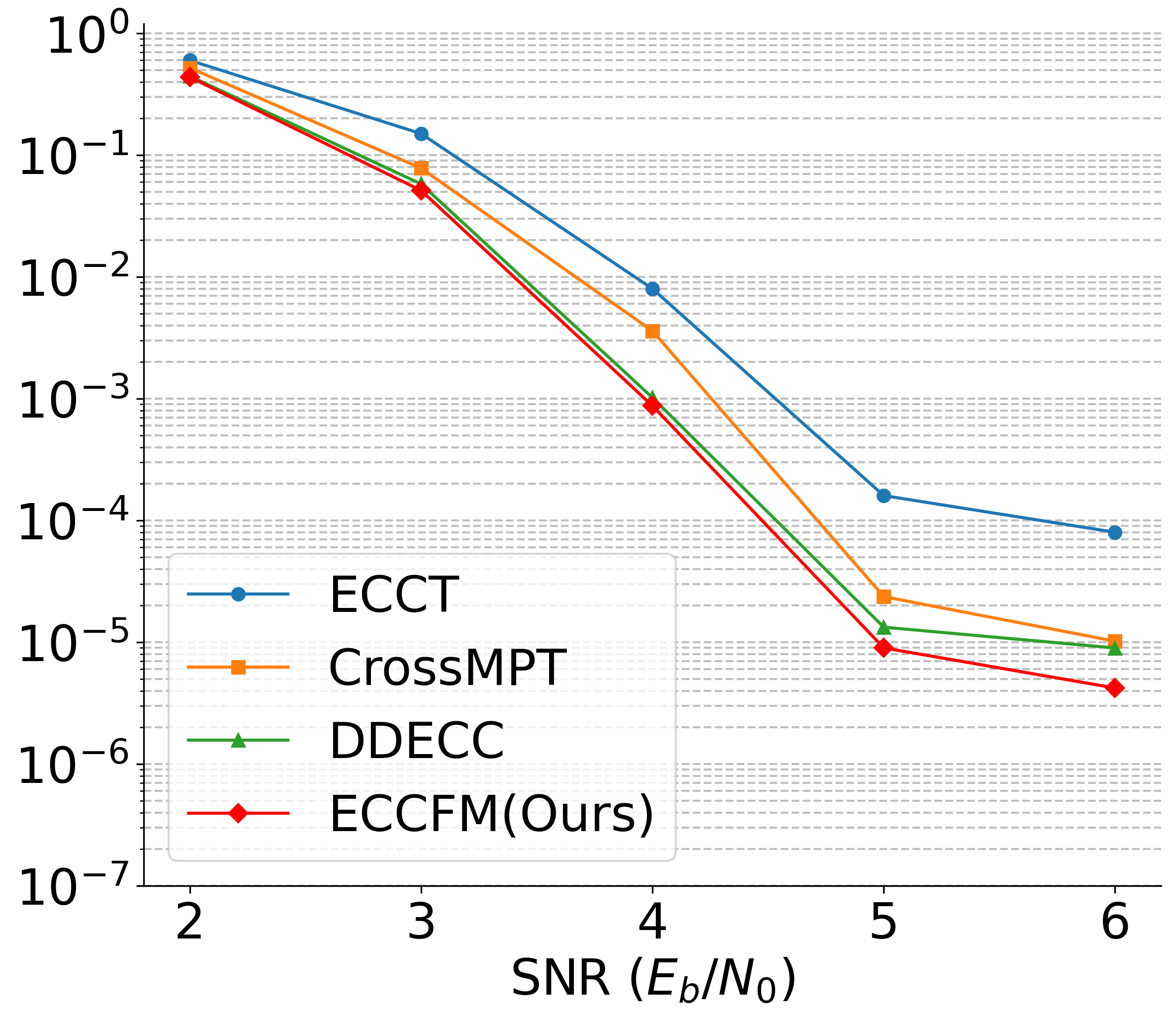}}
    \subfloat[FER on LDPC(408,204)]{\label{fig:b}\includegraphics[width=0.48\textwidth]{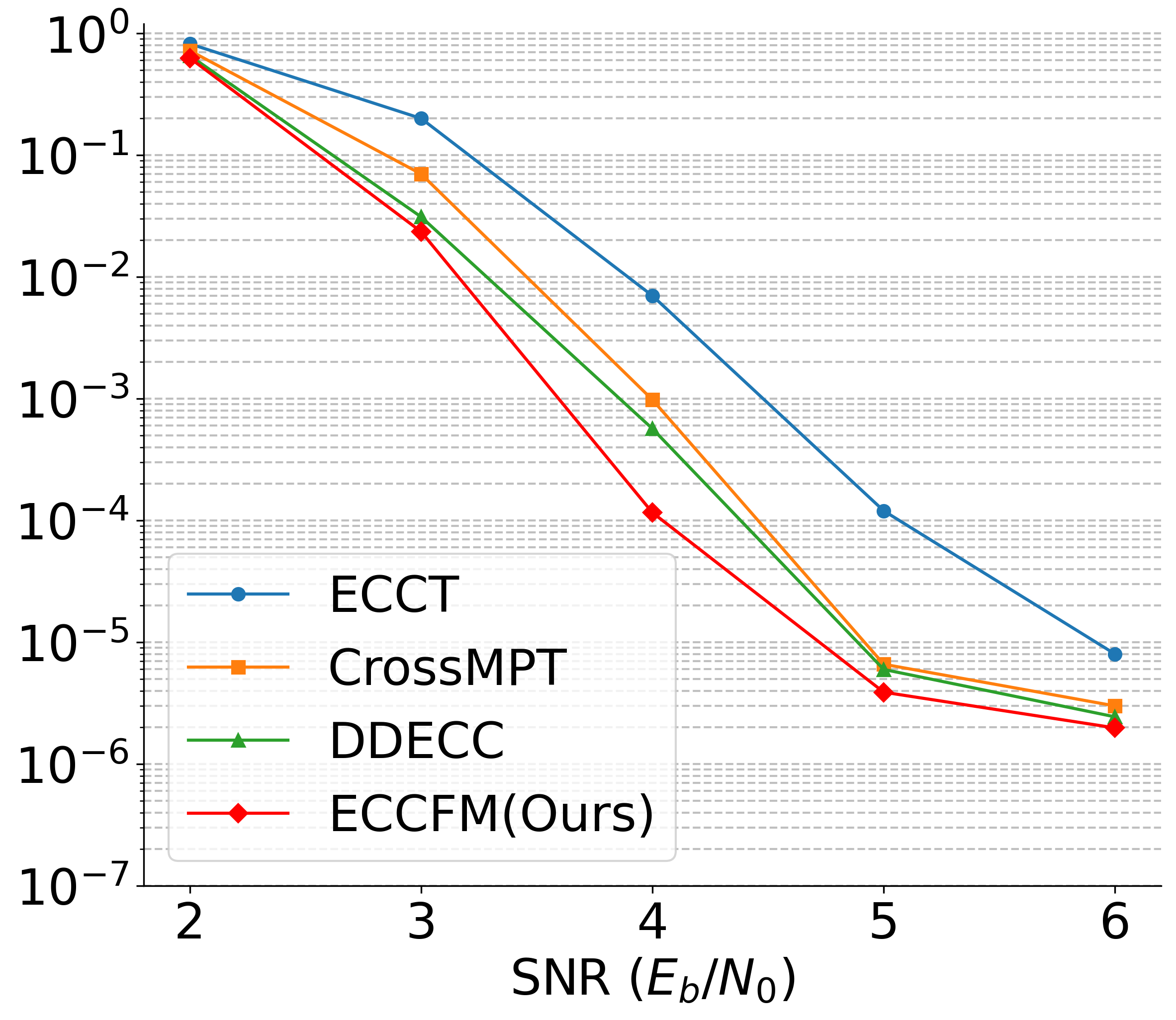}}

    \caption{Performance comparison in terms of Bit Error Rate (BER) and Frame Error Rate (FER) for two LDPC codes with different blocklengths and rates: LDPC($n=204, k=102$) and LDPC($n=408, k=204$). Our method is evaluated against ECCT, CrossMPT, and DDECC.}
    \label{fig:four_images}
\end{figure}

As established in Figure~\ref{fig:longer}, ECCFM demonstrates scalability, achieving competitive performance on both short and medium-to-long codes. To further validate this, we present additional results for LDPC codes of varying lengths and rates, specifically LDPC($n=204, k=102$) and LDPC($n=408, k=204$). The BER and FER results in Figure~\ref{fig:four_images} confirm that ECCFM improves decoding performance while maintaining its high inference speed.

\subsubsection{Additional Results on Inference Time and Throughput} \label{appendix_inference}
\begin{figure}[htbp]
    \centering
    \subfloat[LDPC(121,60)]{\label{fig:a}\includegraphics[width=0.4\textwidth]{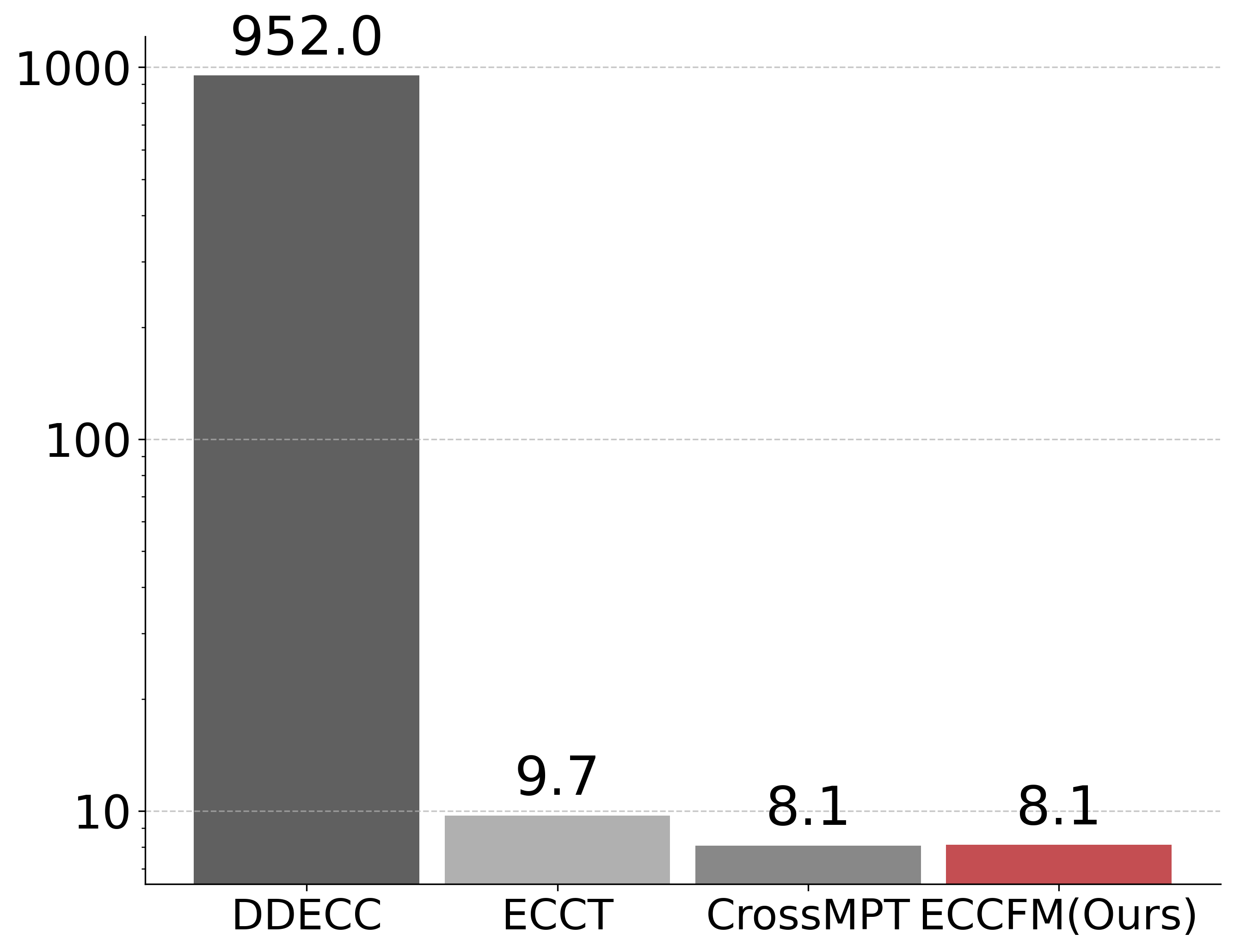}}
    \subfloat[LDPC(121,70)]{\label{fig:b}\includegraphics[width=0.4\textwidth]{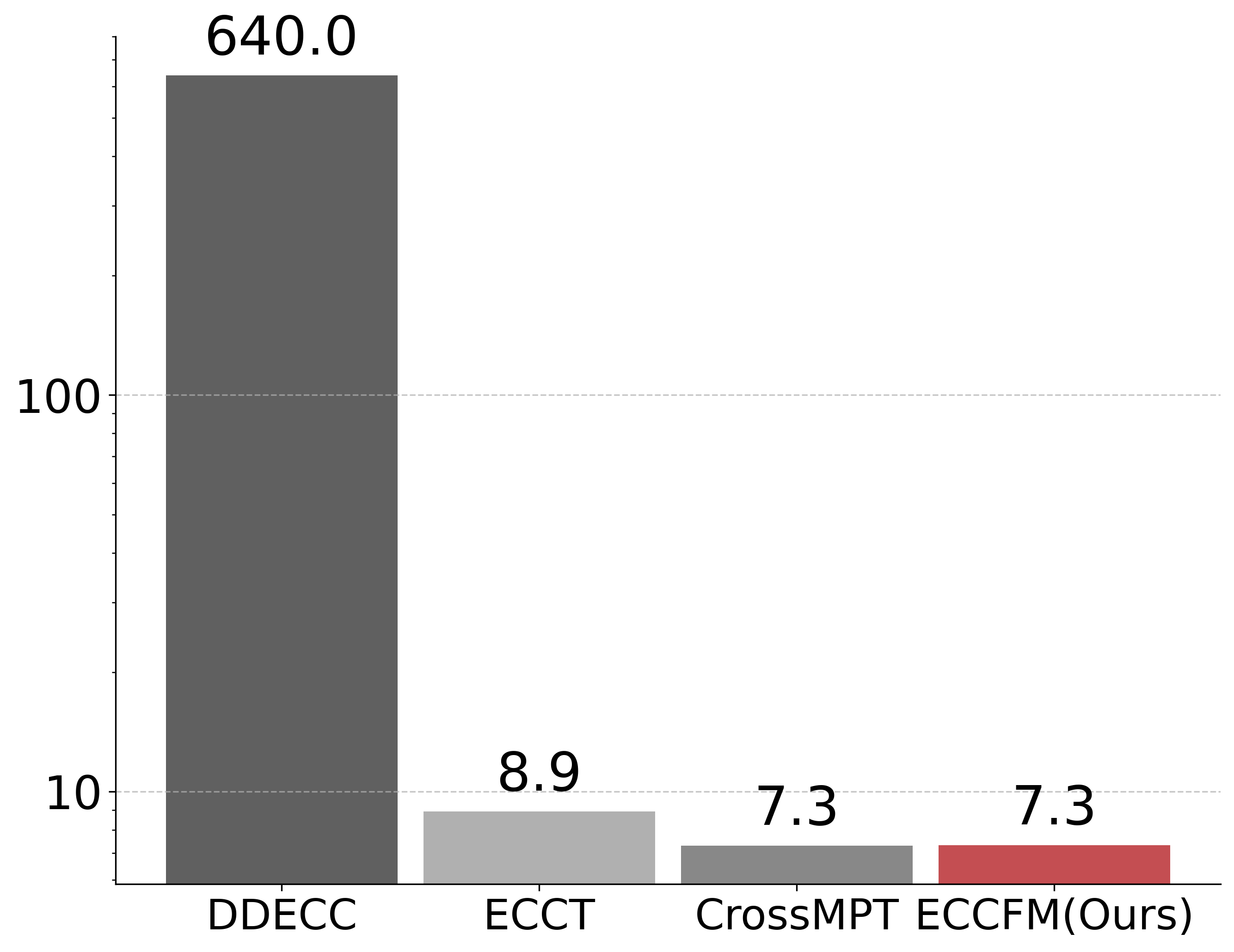}}
    \\[0.2cm]
    \subfloat[LDPC(121,80)]{\label{fig:b}\includegraphics[width=0.4\textwidth]{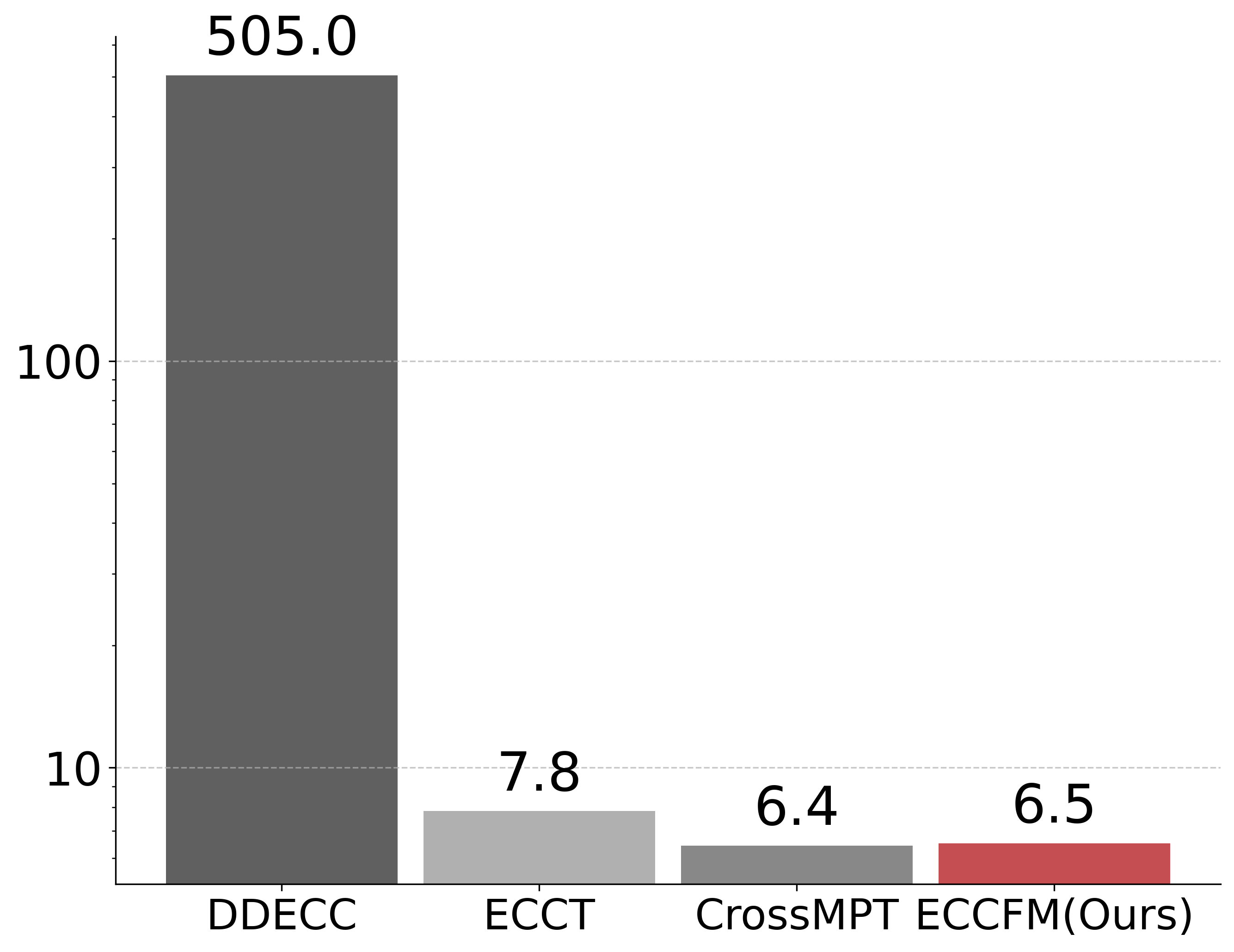}}
    \subfloat[LDPC(529,440)]{\label{fig:b}\includegraphics[width=0.4\textwidth]{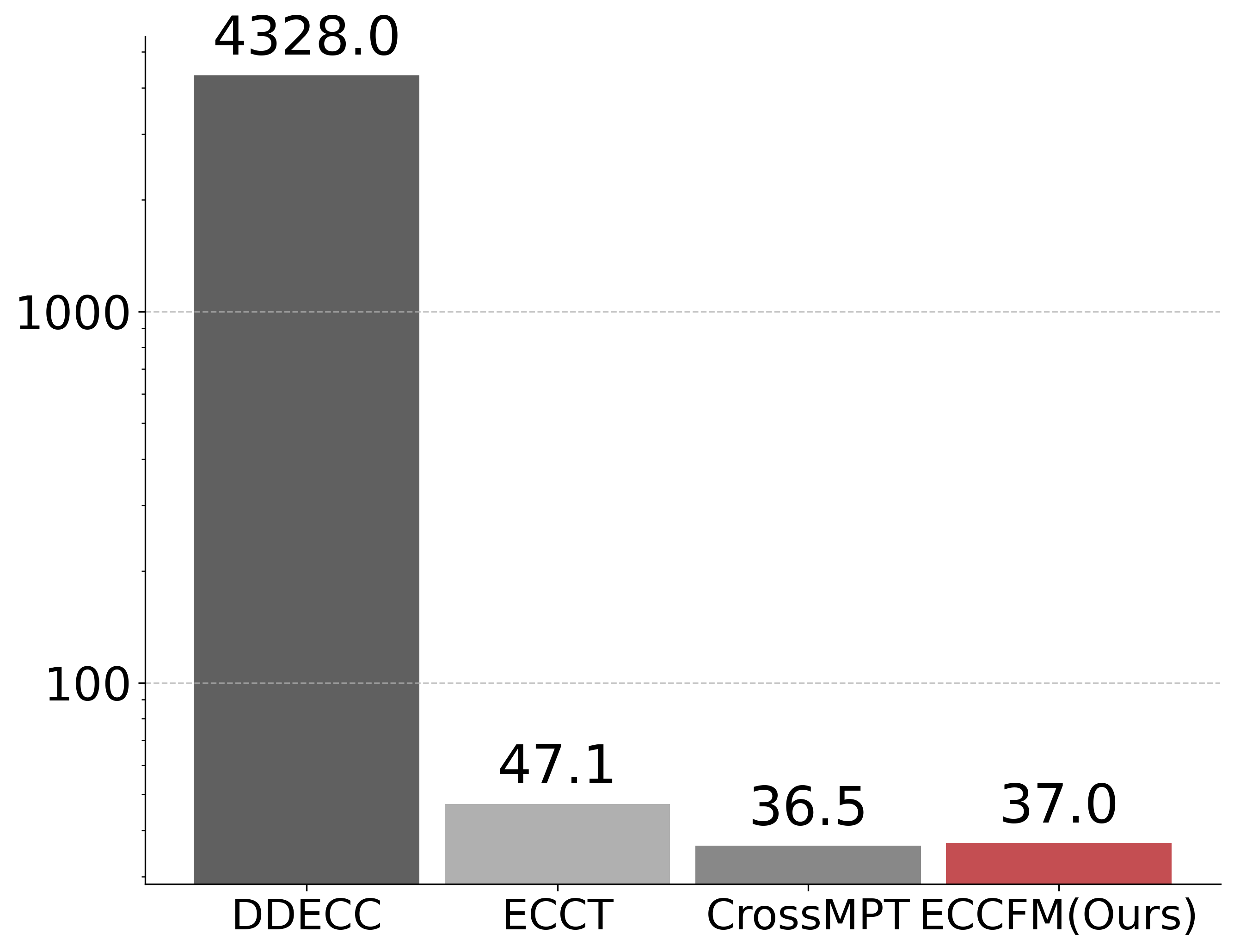}}

    \caption{Inference time on LDPC($n=121,k=60$), LDPC($n=121,k=70$), LDPC($n=121,k=80$) and LDPC($n=529,k=440$), comparing with ECCT, CrossMPT and DDECC.}
    \label{fig:additional_inference_ldpc}

    \subfloat[LDPC(121,60)]{\label{fig:a}\includegraphics[width=0.4\textwidth]{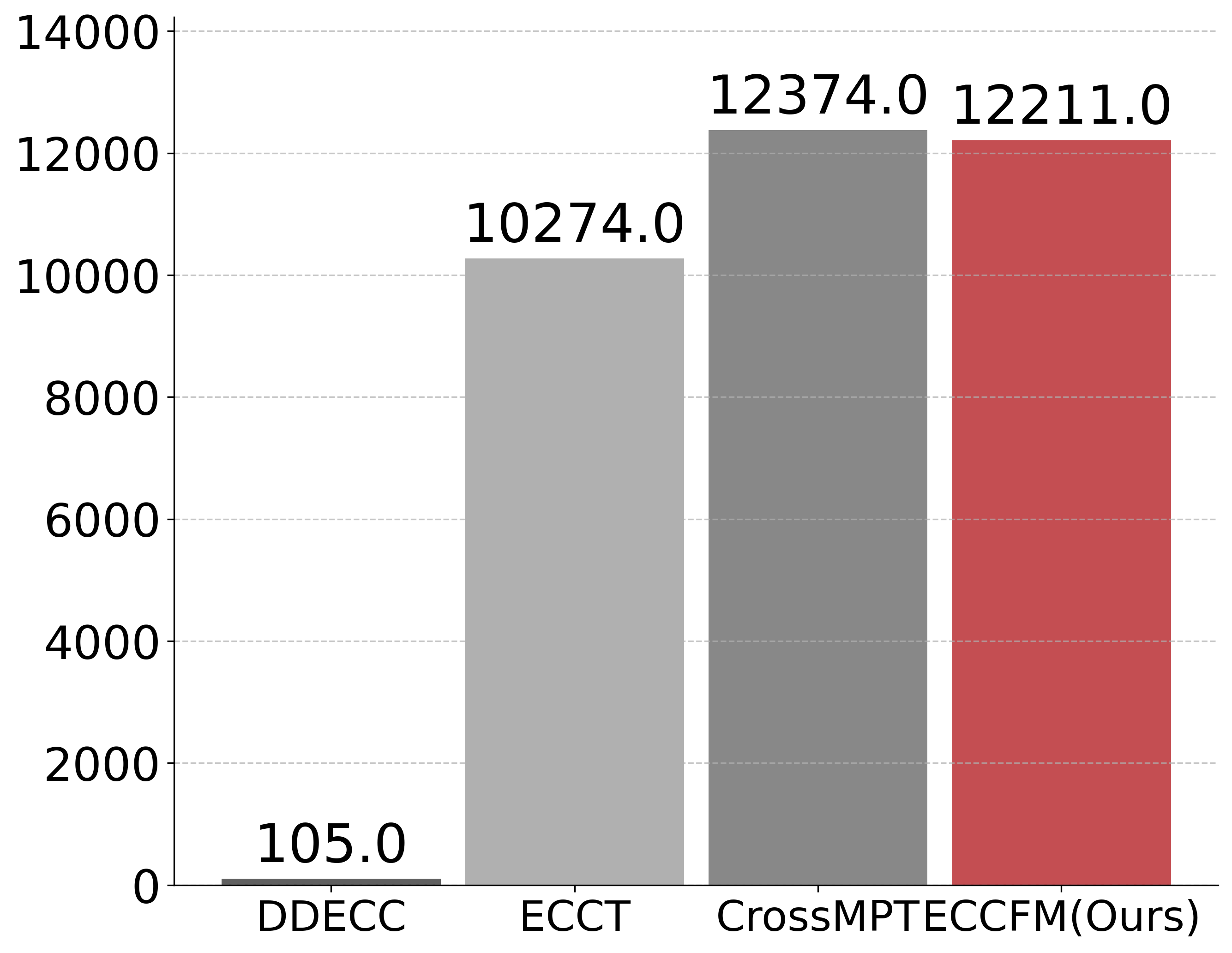}}
    \subfloat[LDPC(121,70)]{\label{fig:b}\includegraphics[width=0.4\textwidth]{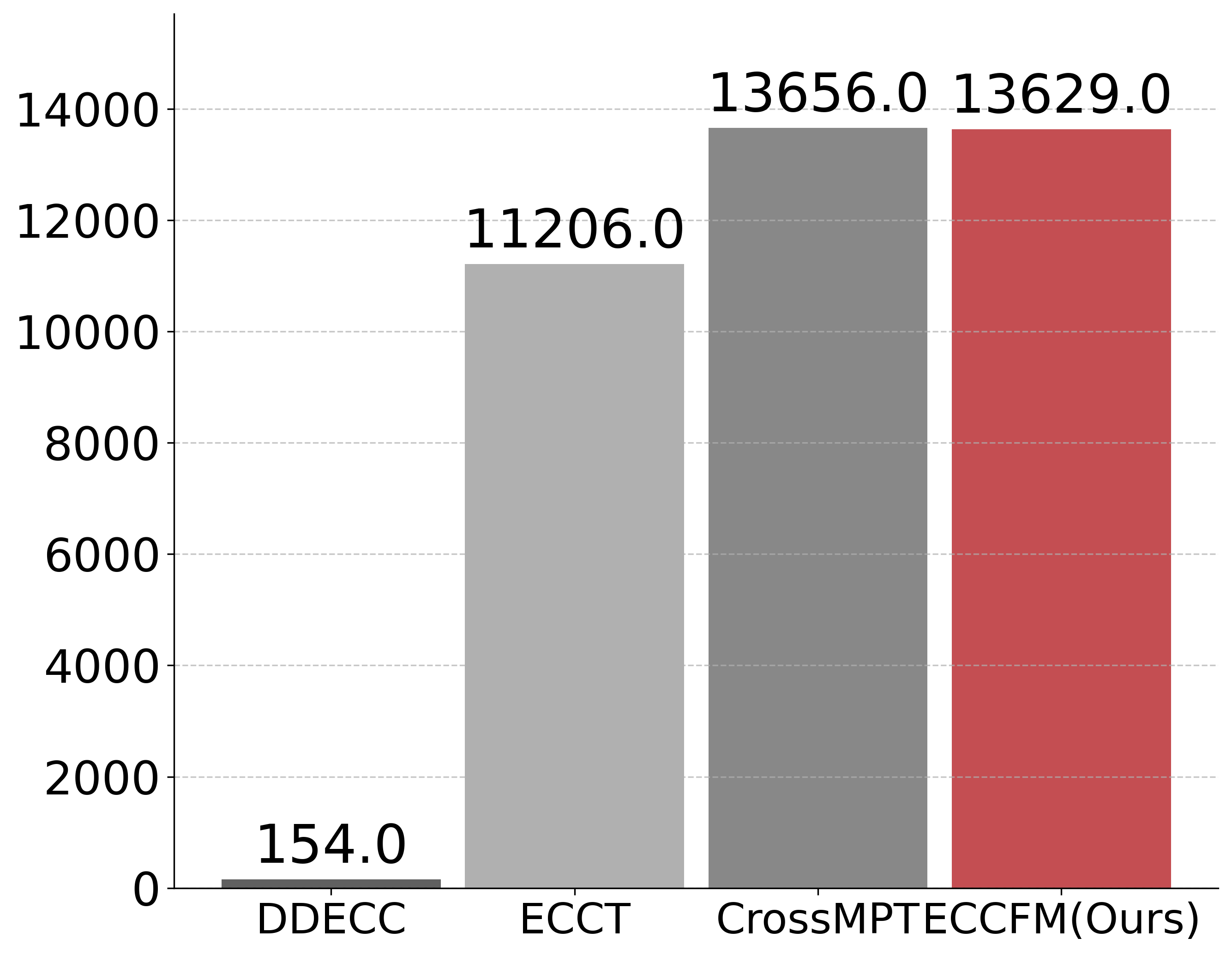}}
    \\[0.2cm]
    \subfloat[LDPC(121,80)]{\label{fig:b}\includegraphics[width=0.4\textwidth]{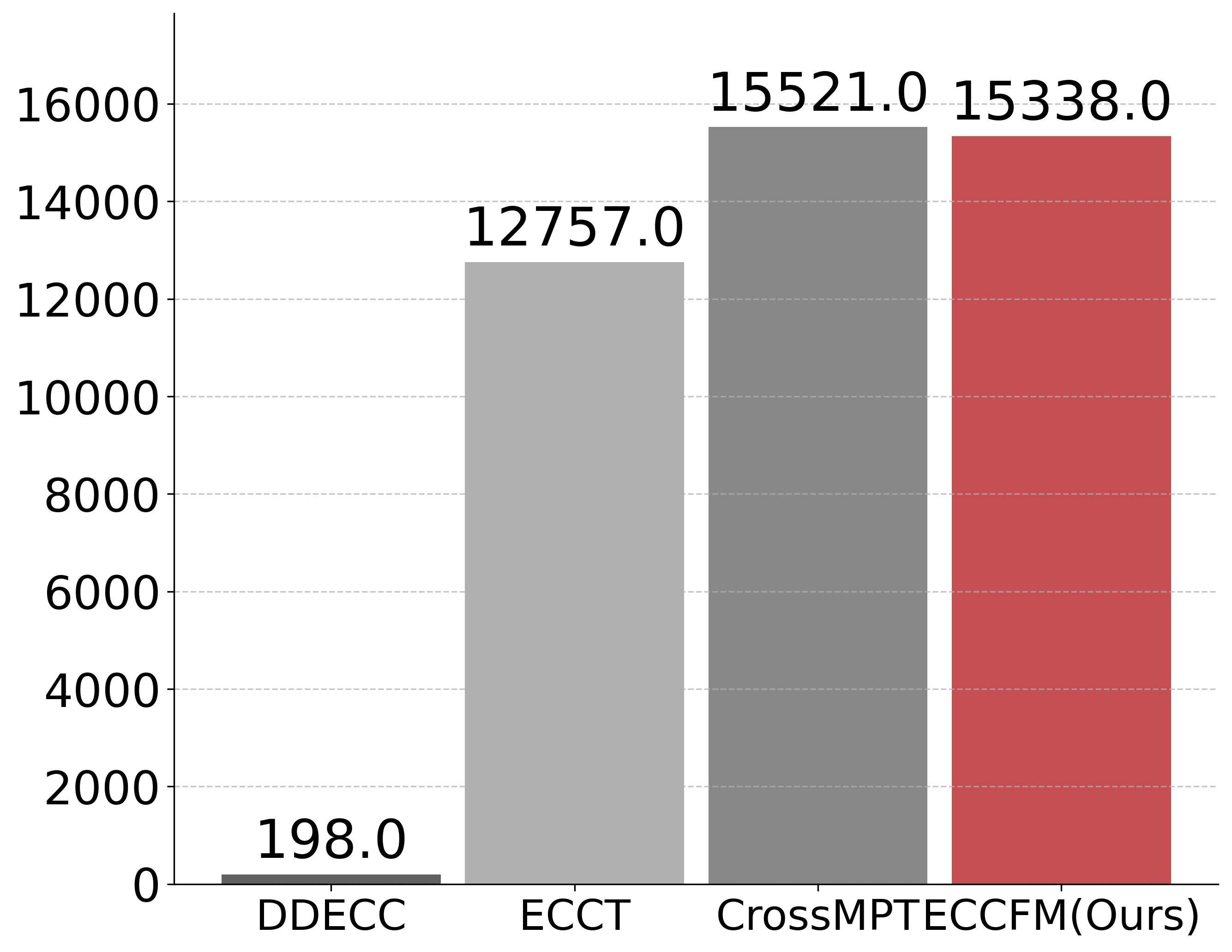}}
    \subfloat[LDPC(529,440)]{\label{fig:b}\includegraphics[width=0.4\textwidth]{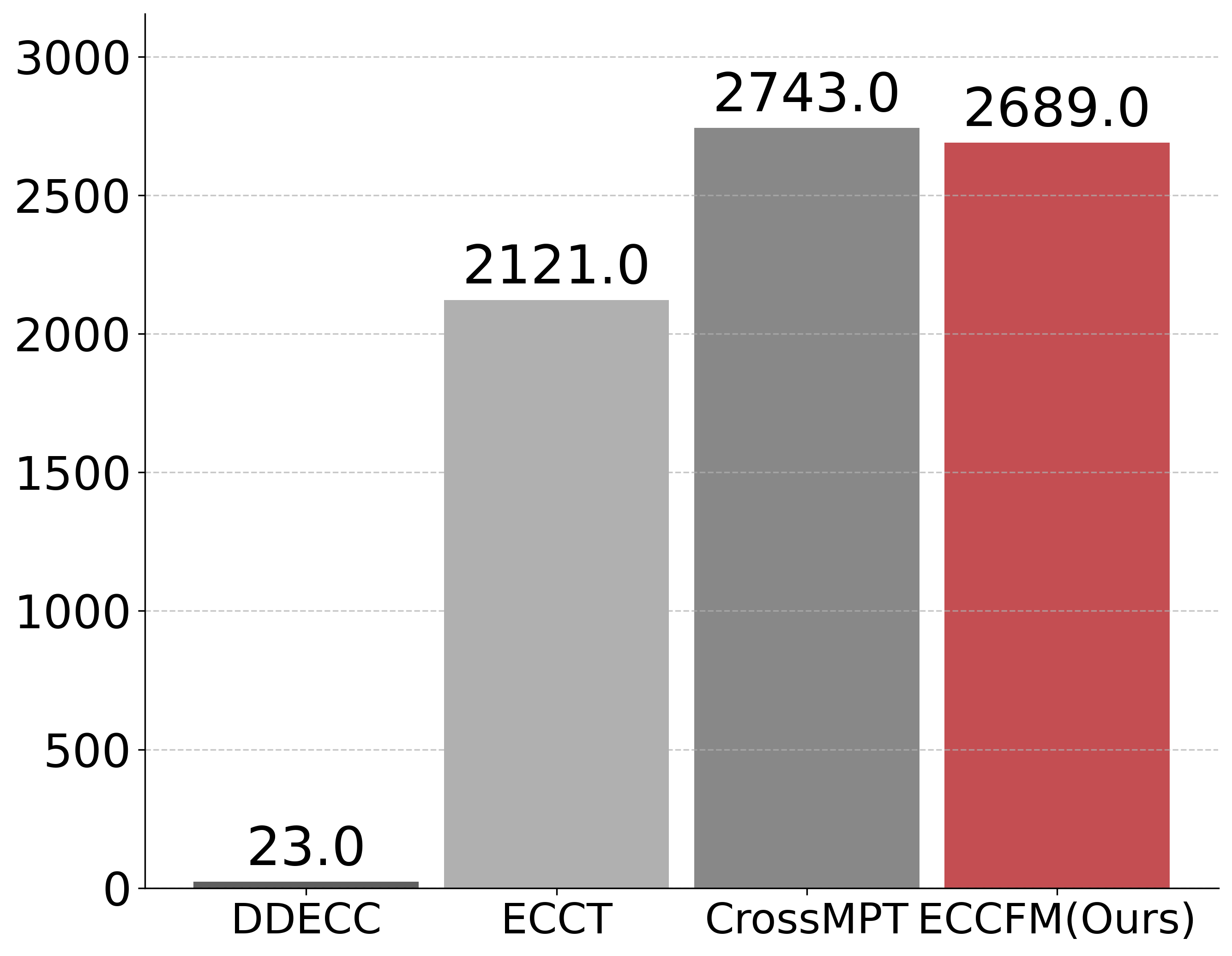}}

    \caption{Throughput on LDPC($n=121,k=60$), LDPC($n=121,k=70$), LDPC($n=121,k=80$) and LDPC($n=529,k=440$).}
    \label{fig:additional_throughput_ldpc}
\end{figure}

\begin{figure}[htbp]
    \centering
    \subfloat[POLAR(128,86)]{\label{fig:a}\includegraphics[width=0.32\textwidth]{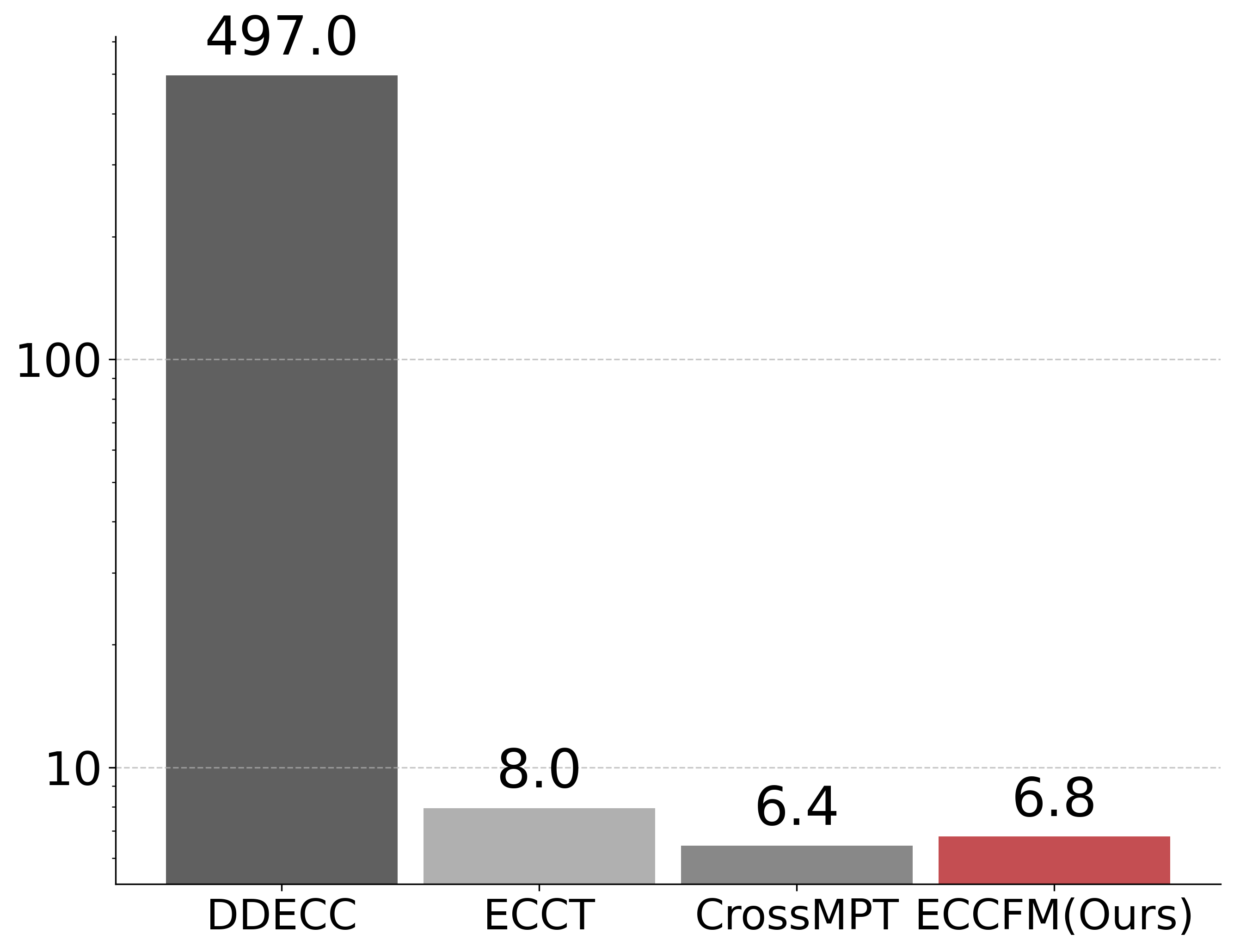}}
    \subfloat[POLAR(128,96)]{\label{fig:b}\includegraphics[width=0.32\textwidth]{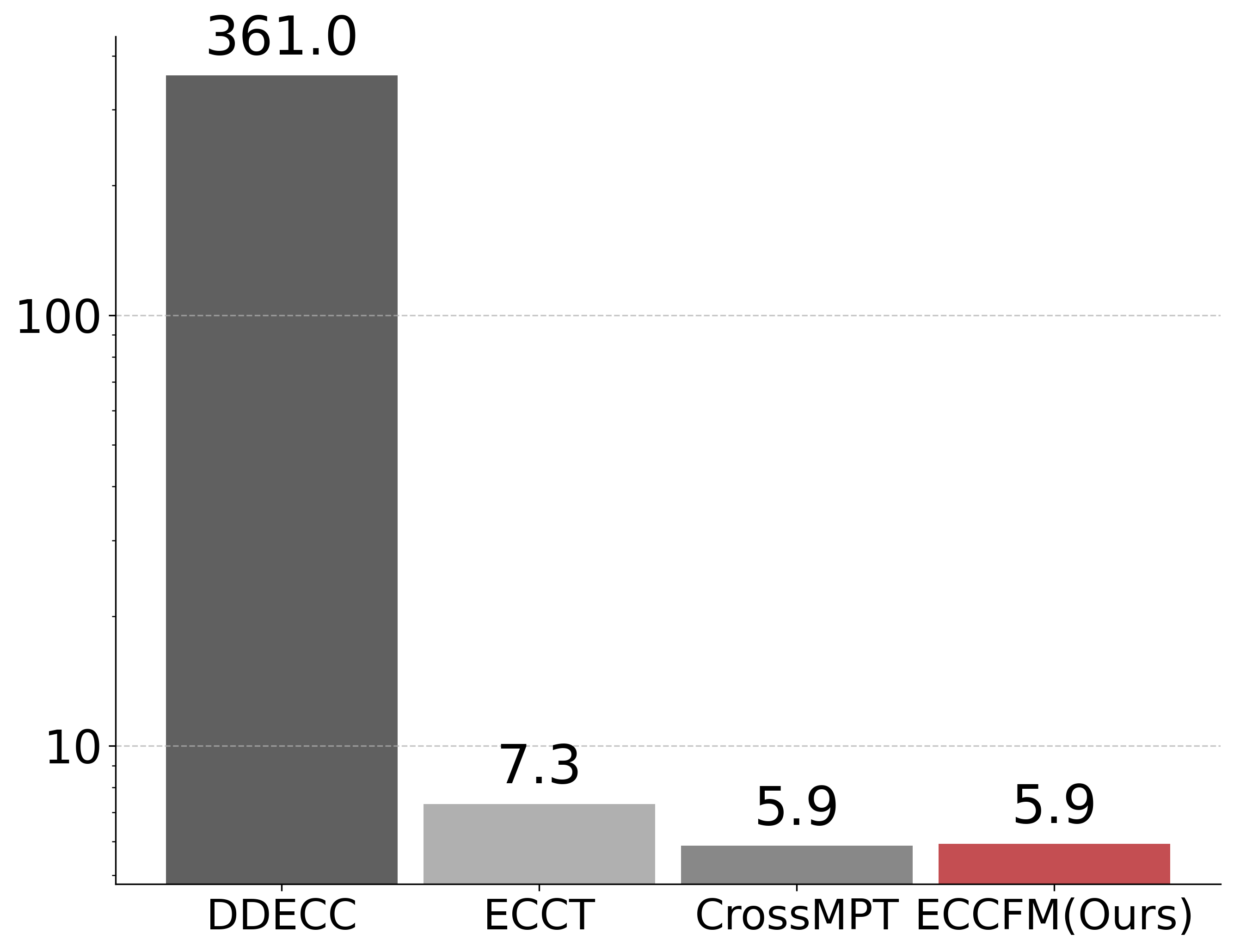}}
    \subfloat[POLAR(512,384)]{\label{fig:b}\includegraphics[width=0.32\textwidth]{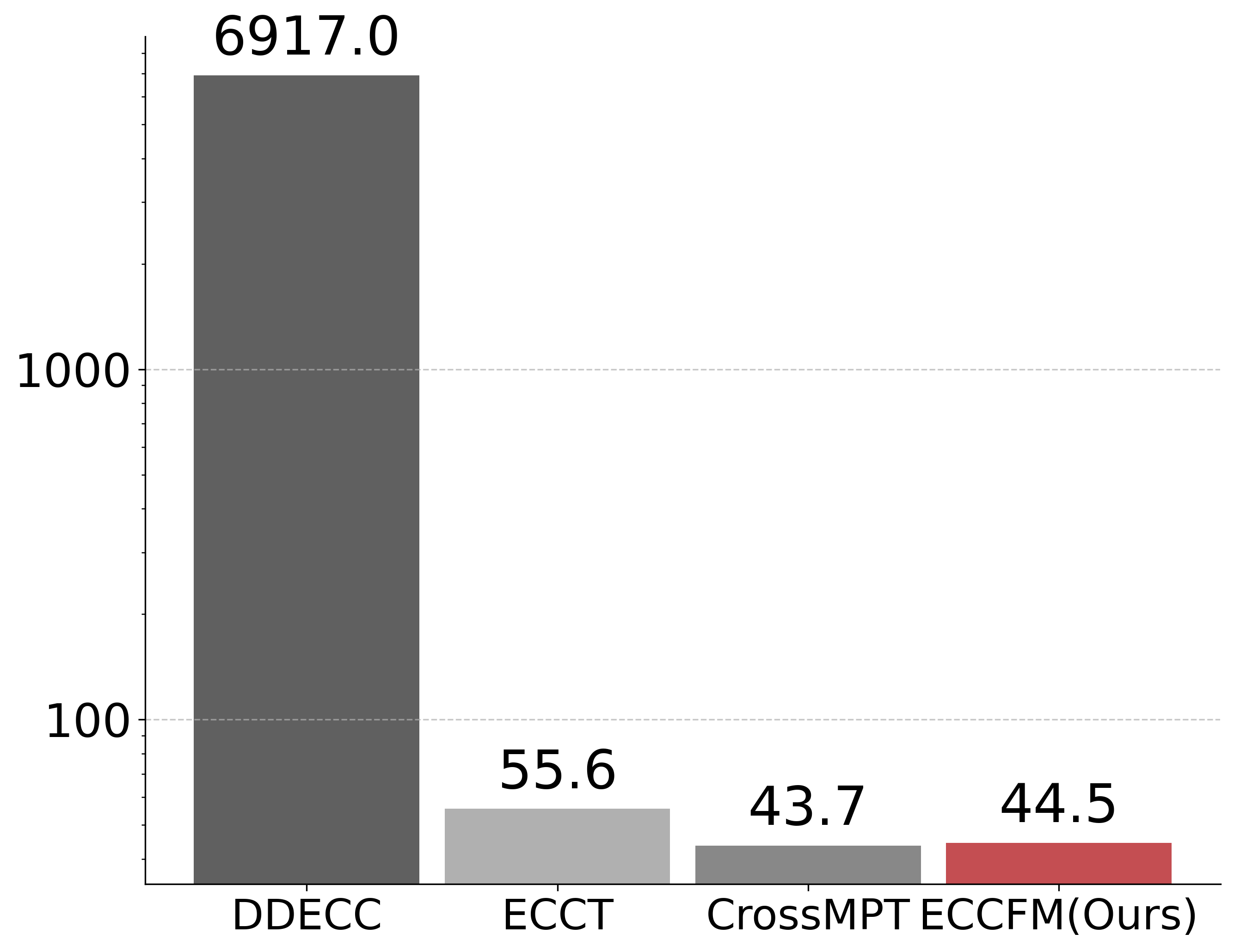}}
    \caption{Inference time on POLAR($n=128,k=86$), POLAR($n=128,k=96$) and POLAR($n=512,k=384$), comparing with ECCT, CrossMPT and DDECC.}
    \label{fig:additional_inference_polar}

    \subfloat[POLAR(128,86)]{\label{fig:a}\includegraphics[width=0.32\textwidth]{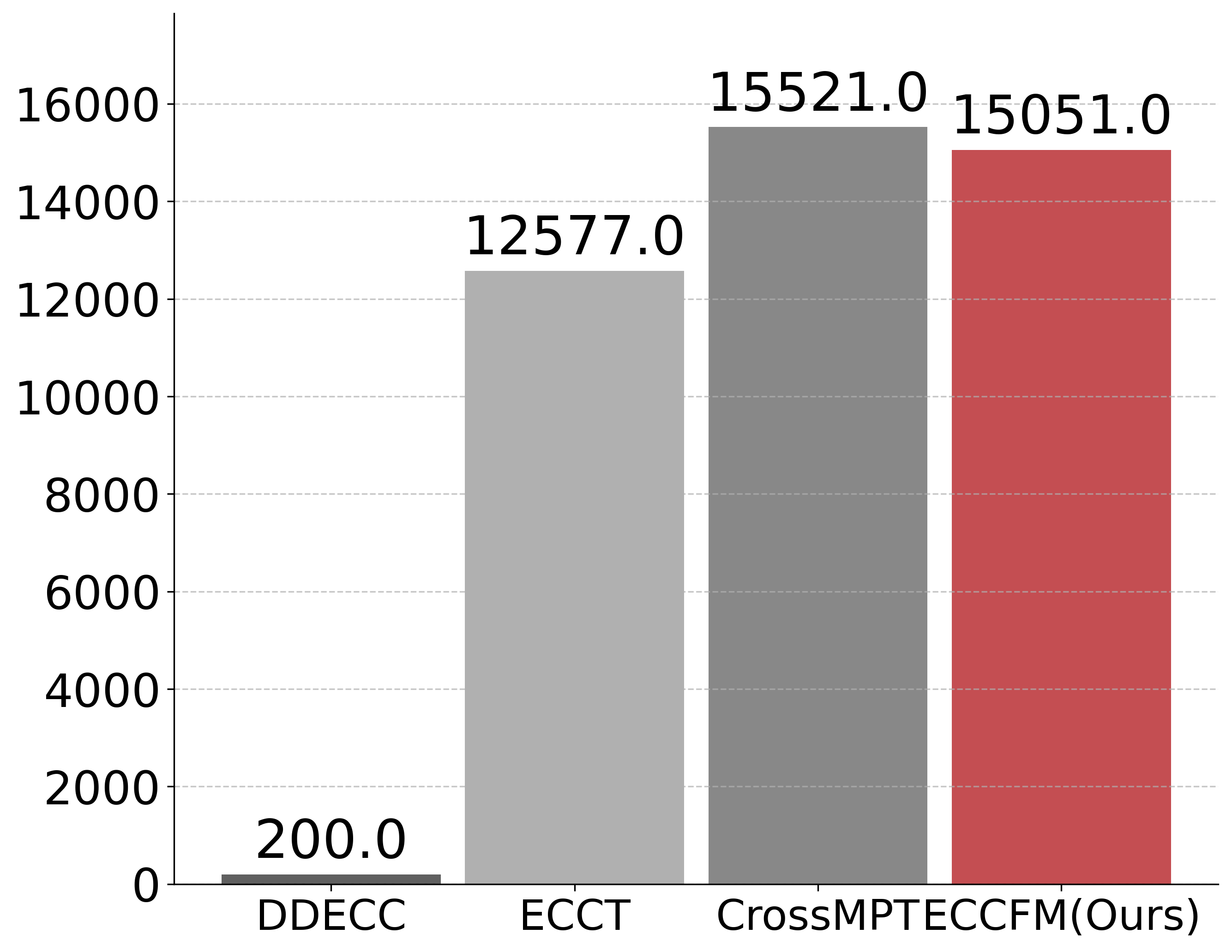}}
    \subfloat[POLAR(128,96)]{\label{fig:b}\includegraphics[width=0.32\textwidth]{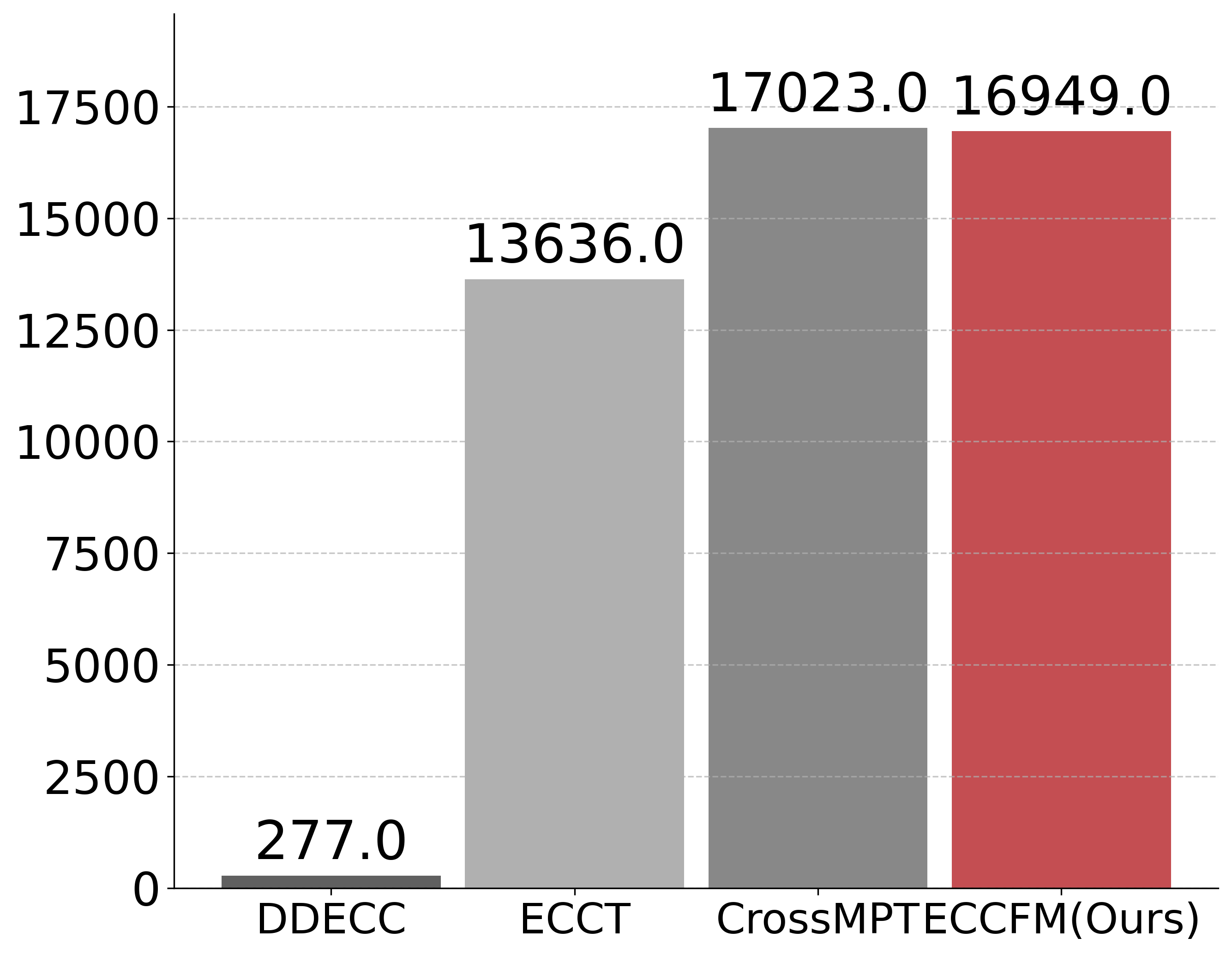}}
    \subfloat[POLAR(512,384)]{\label{fig:b}\includegraphics[width=0.32\textwidth]{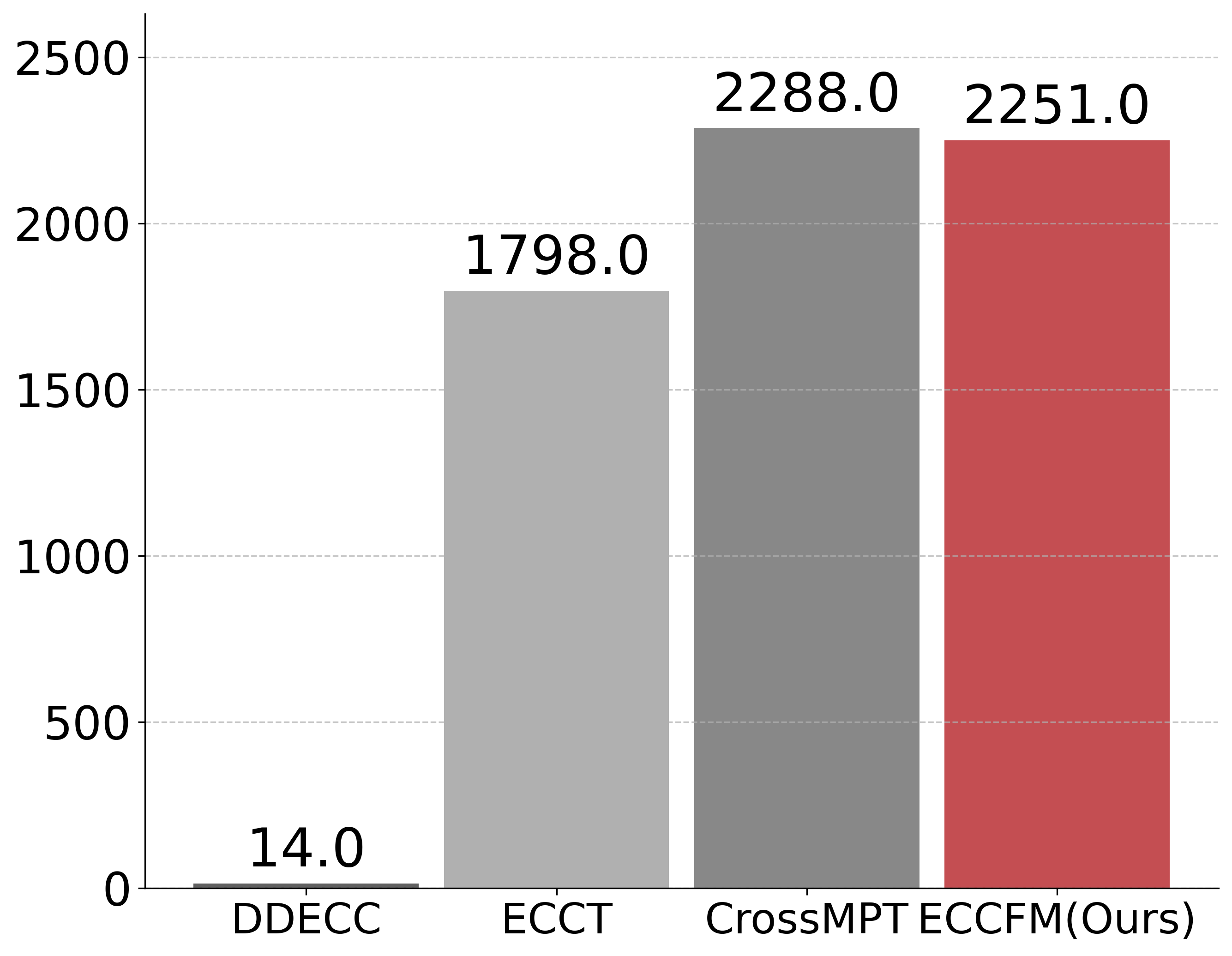}}
    \caption{Inference time on POLAR($n=128,k=86$), POLAR($n=128,k=96$) and POLAR($n=512,k=384$).}
    \label{fig:additional_throughput_polar}
\end{figure}

To evaluate the practical efficiency of ECCFM, we benchmarked its inference time and throughput on both POLAR and LDPC codes against established baselines (ECCT, CrossMPT, and DDECC). Inference time was measured as the total duration in seconds to decode $10^6$ samples, while throughput was defined as the number of samples decoded per second. As illustrated in Figure~\ref{fig:inference}, ECCFM achieves a speedup over the denoising diffusion method, DDECC. The advantage scales with code complexity, growing from a 30x speedup on short codes to over 100x on longer codes. Further analysis, detailed in Figure~\ref{fig:additional_inference_ldpc}, Figure~\ref{fig:additional_throughput_ldpc}, Figure~\ref{fig:additional_inference_polar}, and Figure~\ref{fig:additional_throughput_polar} confirms that these efficiency gains are consistent across a wide range of code types, lengths, and rates. This performance demonstrates that ECCFM provides a significant improvement in decoding speed, particularly for long codes where latency is a critical bottleneck.

\subsubsection{Analysis on Computational Overhead of iterative denoising phase}
To point out the ECCFM's speed advantage, we analyzed the iterative convergence of the denoising diffusion framework. Specifically, we measured the average number of inference steps required for the DDECC model to converge to a valid codeword (i.e., achieve a zero syndrome, $e_t=0$). As detailed in Table~\ref{tab:converge}, the computational overhead for DDECC increases substantially under these three conditions: longer codes, lower code rates, and lower SNRs. This is precisely the bottleneck that ECCFM's one-step decoding offers a consistent gain in efficiency, particularly in these difficult decoding scenarios.

\begin{table*}[h]
    \centering
    \caption{Convergence steps to $e_t=0$ of the DDECC decoder on longer codes across different Signal-to-Noise Ratios ($E_b/N_0$). The results are reported in terms of the average steps(variance).}
    \label{tab:converge}
    \renewcommand{\arraystretch}{1.25}
    \resizebox{0.95\textwidth}{!}{
        \begin{tabular}{cc ccccc}
        \toprule
        \multirow{2}{*}{\textbf{Code Type}} & \multirow{2}{*}{\textbf{Parameters}} & \multicolumn{5}{c}{\textbf{Converge Steps: Average(Variance)}} \\
        \cmidrule(lr){3-7}
        & & \textbf{2} & \textbf{3} & \textbf{4} & \textbf{5} & \textbf{6}\\
        \midrule
        \textbf{POLAR}
        & (512,384) & 123.40(11.38) & 91.34(21.68) & 60.24(18.00) & 41.40(16.13) & 24.99(14.82) \\ 
        \midrule
        \multirow{3}{*}{\textbf{LDPC}} 
        & (204,102) & 57.10(21.51) & 39.37(10.50) & 29.47(7.50) & 21.25(6.90) & 14.24(6.31) \\
        & (408,204) & 112.12(39.47) & 76.91(13.18) & 58.39(10.38) & 42.39(9.62) & 28.99(8.81) \\
        & (529,440) & 89.54(8.07) & 59.25(27.75) & 27.48(9.78) & 17.06(6.95) & 9.22(6.03) \\
        \midrule
        \textbf{WRAN} & (384,320) & 59.66(9.79) & 37.59(18.88) & 18.57(7.99) & 11.23(5.49) & 5.91(4.64) \\ 
        \bottomrule
        \end{tabular}%
    }
\end{table*}

\end{document}